\documentclass{article}
\usepackage{frExamplee}
\usepackage{graphicx}
\usepackage{apalike}
\usepackage{setspace}

\usepackage[dvipsnames]{xcolor}
\usepackage{amsmath}
\usepackage{ulem}
\newcommand{\control}{\textbf{Control}}
\newcommand{\ekfpl}{\textbf{Row Follower EKF}}
\newcommand{\update}{\textbf{Update}}
\newcommand{\autofarm}{IAF}

\usepackage{graphics} % for pdf, bitmapped graphics files
\usepackage{graphicx} 
\usepackage{gensymb}
\usepackage{subcaption}
\usepackage{multirow}
\usepackage{float}

\title{Multi-Sensor Fusion based Robust Row Following for Compact Agricultural Robots}

\author{
Andres Eduardo Baquero Velasquez\\
Department of Agricultural and Biological Engineering\\
University of Illinois at Urbana-Champaign\\
Urbana, IL 61801 \\
\texttt{andru89@illinois.edu} \\
(Corresponding Author)\\
\And
Vitor Akihiro Hisano Higuti \\
EarthSense Co. \\
Champaign, IL 61820 \\
\texttt{akihiro@eathsense.co}
\And
Mateus Valverde Gasparino\\
University of Illinois at Urbana-Champaign\\
Urbana, IL 61801 \\
\texttt{mvalve2@illinois.edu} \\
\And
Arun Narenthiran Sivakumar\\
University of Illinois at Urbana-Champaign\\
Urbana, IL 61801 \\
\texttt{av7@illinois.edu} \\
\And
Marcelo Becker \\
Department of Mechanical Engineering \\
University of Sao Paulo \\
Sao Carlos, SP, Brazil\\
\texttt{becker@usp.br} \\
\And
Girish Chowdhary\\
Coordinated Science Lab\\
University of Illinois at Urbana-Champaign\\
Urbana, IL 61801 \\
\texttt{girishc@illinois.edu} \\
}

\begin{document}

\maketitle

\begin{abstract}
This paper presents a state-of-the-art LiDAR based autonomous navigation system for under-canopy agricultural robots.  Under-canopy agricultural navigation has been a challenging problem because GNSS and other positioning sensors are prone to significant errors due to  attentuation and multi-path caused by crop leaves and stems. Reactive navigation by detecting crop rows using LiDAR measurements is a better alternative to GPS but suffers from challenges due to occlusion from leaves under the canopy. Our system addresses this challenge by fusing IMU and LiDAR measurements using an Extended Kalman Filter framework on low-cost hardwware. In addition, a local goal generator is introduced to provide locally optimal reference trajectories to the onboard controller. Our system is validated extensively in real-world field environments over a distance of 50.88~km on multiple robots in different field conditions across different locations. We report state-of-the-art distance between intervention results, showing that our system is able to safely navigate without interventions for 386.9~m on average in fields without significant gaps in the crop rows, 56.1~m in production fields and 47.5~m in fields with gaps (space of 1~m without plants in both sides of the row).
\end{abstract}

%Our system works by fusing IMU data with LiDAR measurements and utilizes a Local Goal Generator for robust navigation of compact robots in cluttered under-canopy environments.

\section{Introduction}

Robotics and digital agriculture technologies hold great potential in improving the sustainability, productivity, and access to agriculture \cite{sparrow2020robots}. Labor remains one of the key issues facing sustainable intensification. As a result, many research groups around the world  have focused their efforts on automating various agricultural tasks, such as planting and harvesting, plant scouting, detection and treatment of pests and diseases \cite{Nevesrobots2017,oliveira2022Robotics,shamshiri2018Research}. The expectation is that intelligent automation can provide ways to mitigate the impact of major global problems such as climate change, soil depletion, loss of biodiversity, water scarcity, and population growth \cite{sparrow2020robots}. 

However, automating existing large equipment alone may not enable the future that many envision. Large equipment can cause soil compaction, are expensive to run and transport, and provide significant safety hazards for autonomy. Farm sizes, and accordingly machines, have kept becoming larger \cite{Chen2018,king2017future} to the point where some authorities have indicated that the agricultural mechanization of a country should be measured as instrument/machine weight per tractor (kg/tractor), toll/machine number per tractor, among others \cite{akdemir2013agricultural}. However, this sole focus on ``bigger is better'' is limiting, and stems mainly from the labor bottleneck that most agricultural tasks face. As robotics and sensing technologies improve, they can provide novel solutions that could be more efficient and cost-effective than large machines \cite{king2017future}. These small robots could also make agriculture profitable at \textit{small} as well as large scale, helping improve agricultural diversity, and productivity in developing nations.

Such agricultural robots (or Agbots) may help farmers to improve yield and productivity while reducing levels of fertilizer and pesticide as well as water wastage \cite{sparrow2020robots}. Indeed, topsoil compaction has been associated with reduced agricultural productivity. In this context, lighter (weight less than 33 tons) teleoperated or autonomous robots potentially reduce the compaction issue \cite{sparrow2020robots,fue2020extensive}. Small autonomous under-canopy robots that can travel in between the rows of crops could provide new tools to growers of commodity crops such as corn (\textit{Zea mays}), soybean (\textit{Glycine max}), and cotton (\textit{Gossypium}). For example, small under-canopy robots can enable mechanical weeding \cite{Bakker2006weeding,Reiser2019weeding,Wyatt2019Agbots,Wyatt2020Agbots2}, cover-crop planting \cite{cavender-bares_2021}, and under-canopy high-throughput phenotyping \cite{Kayacan2018EmbeddedRobot,Mueller-Sim2017,Vinobot}. This has led to significant interest in autonomous small robots for agriculture which have become focus of study for various research groups around the world. There are also several companies focused on different form factors, including Naio, Ecorobotix, Row-Bot, and EarthSense, which developed the TerraSentia robot that is used in this study. \cite{RaminShamshiri2018ResearchFarming} and \cite{Fountas2020AgriculturalOperations} provide a systematic review of various applications of agricultural robots and list different research and commercial agricultural robotic platforms developed and used in crop field operations.

However, reliable autonomy remains the key challenge preventing widespread adoption of small robots. Although there are a significant number of publications about the navigation systems for Agbots using different sensors as GPS, LiDAR or cameras, they are predominantly focused on over-the-canopy robots. Among the limited work that exists in under-canopy navigation, extensive field validation in diverse real-world conditions has not been performed and no other work has shown an extensive autonomous navigation as we show in this paper for this type of environment.  This paper presents significant field results demonstrating that sensor fusion with LiDAR and IMU can enable reliable under-canopy autonomy. GPS is often unreliable under plant canopy and near to the ground due to multi-path and signal attenuation effects. Furthermore, RTK correction is often observed to be not always available, especially in adverse weather and in remote locations. Although cameras are commonly used in the robot navigation in indoor environments, they are more sensitive to illumination conditions and also present shorter range of view compared to LiDAR \cite{Hiremath2014,Reina2016}. Both types of sensor (LiDAR and cameras) are affected by weather phenomena (e.g. fog and rain) or other environmental factors (e.g. dust, smoke and occlusion), the LiDAR is able to provide off-the-shelf distance measurements with accurate ranging over medium range (up to 30$\sim$40~m) and fast operation \cite{Reina2016}. 

In this paper, we show that fusing IMU data with LiDAR measurements and by utilizing a Local Goal Generator to create feasible paths significantly increase the distance between intervention without increasing the cost of the overall system. We report 50.88~km of an  autonomous  solution  for  under-canopy  navigation  in  corn crops where we got an average distance between interventions of 386.9~m for fields without gaps (IAF and ES \# 2), 56.1~m for production fields and 47.5~m for fields (ES \# 1) with gaps (space of 1~m without plants in both sides of the row). Additionally we show that the use of EKF to improve the distances and robot heading estimated by the perception system helps to increase the distances between intervention. It went from  51.6~m without EKF (PL)  to 400~m with EKF (PL+EKF) in fields without gaps, and from 16.3~m without EKF to 56.1~m with EKF in the production fields. Our results show that the proposed methods lead to significantly improved autonomy than earlier single-sensor (only LiDAR based attempts). \cite{higuti2019under} demonstrated that low-cost 2-D LiDAR can be a viable option for navigation in under-canopy environments. They also utilized distance between interventions as a key metric to evaluate reliability of autonomous navigation in a variety of crop environments. However, the reported results in \cite{higuti2019under} were limited to 41.4~m. On the other hand, our results are highly promising, as they indicate that low-cost LiDAR based navigation can be a reliable and effective means of under-canopy navigation.

\section{Related Works}

Autonomous navigation of agricultural robots in general has been studied for a long time, especially with tractors. \cite{RaminShamshiri2018ResearchFarming} and \cite{Fountas2020AgriculturalOperations} provide a systematic review of various applications of agricultural robots and list different research and commercial agricultural robotic platforms developed and used in crop field operations.

Earlier work on autonomous navigation in agriculture was focused on auto guidance of large agricultural machinery. GNSS-based navigation systems have been widely used to allow automatic guidance of traditional crop machinery such as harvesters \cite{Stoll2001,Pilarski2002} and tractors \cite{reid2000agricultural,Bell2000,Thuilot2002,Lenain2003,Abidine2004AutoguidanceDamage,Blackmore2004,rev_Norremark2008,Zhang2016,Wang2016}. In addition, researchers  have also investigated the potential of using GNSS based navigation systems for small autonomous robots that navigate over-the-canopy on horticultural crops such as sugar beet \cite{Bak2004,Bakker2011}. But GNSS based navigation suffers from inaccuracies due to multi path error from plant canopies in case of under-canopy navigation in row crops such as corn, sorghum etc. Robatanist, an under-canopy robotic platform developed for sorghum phenotyping is an example \cite{Mueller-Sim2017}. It uses GNSS for navigation and hence faces challenges when the sorghum plants grow taller than the level of the GNSS antenna. However, under dense crop canopies, the GNSS readings deteriorate due to occlusion, attenuation, and multi-path errors. A reported example is Robotanist, a thin robot capable of driving through a single lane to automate the phenotyping process in sorghum crops . The authors highlight that GNSS navigation is a capability to be lost as season progresses and sorghum plants becomes taller than GNSS antenna. Also, GNSS is not reliable in some countries in the southern hemisphere because of the South Atlantic Magnetic Anomaly (SAMA) that affects the GNSS signal \cite{Abdu2005SouthIonosphere,Spogli2013AssessingActivity}. Additionally, since GPS is an extrinsic sensor that only provides the position of the robot, it cannot overcome challenges such as presence of static and dynamic obstacles such as rocks, animals etc in the planned GNSS waypoint path \cite{Reina2016,Rovira-Mas2015}. This shows a GNSS based navigation cannot be a stand alone solution for autonomous agricultural robots. In particular, GNSS is not reliable for autonomous navigation in orchards and under-canopy environments \cite{Bergerman2015,Santos2015,Mueller-Sim2017}. This led to work on agricultural navigation using vision and LiDAR based systems.

Vision based navigation systems have been extensively studied for the row following problem in agricultural fields. The geometric structure of the crop rows as parallel lines in agricultural fields enables the use of computer vision. As in the case of GNSS systems, earlier works focused on auto guidance for tractors and heavy machinery such as harvesters \cite{reid1987vision}. Subsequent work also focused on vision based navigation of over-the-canopy agricultural robots \cite{Montalvo2012,Jiang2015,Zhao2016,Zhai2016,ball2016vision}. The camera looks down at the crop rows from a top down view in these cases and a clear view of multiple crop rows as vanishing lines are visible in the images. Therefore the common approach is segmentation of vegetation from the soil background using color indices and line fitting on these segmented images using different approaches such as Hough Transforms. From the fitted lines, the relative orientation and offset of the robot with respect to the crop row can be obtained and used to steer the robot to the middle of the lane between the crop rows \cite{Zhang2018TractorVision,Radcliffe2018MachineNavigation,Hiremath2014a,Subramanian2006,English2014}. Also, sensor fusion with GPS was done in a lot of these cases to improve the robustness of the system. But this approach faces challenges in under-canopy environments in row crops because of the presence of significantly more clutter and occlusion. \cite{Xue2012VariableRobot} developed and tested a vision based navigation system for under-canopy navigation in corn but its validation was limited to cases when the occlusion and clutter is less and the vegetation and soil are distinguishable. In general, this classical vision based navigation approach is affected by continuously changing lighting conditions and appearance of crops and soil in the field throughout the season.

LiDAR sensors are active sensors that sense geometric information of the environment. It is commonly used for mobile robot navigation (localization and obstacle detection) in different applications \cite{siegwart2011introduction,Pouliot2012,Reina2016}. In the context of agricultural navigation, LiDAR has been widely studied mostly in orchard environments \cite{Barawid2007,Zhang2013,Bergerman2015,Bell2016,Lemos2018UNISENSORYLASER}. They use 2D LiDAR sensors and similar line fitting on images to detect crop rows, lines are fit on the 2D laser scans to detect the orientation and offset of the robot with respect to the crop rows. Some work also studied the use of 3D LiDAR sensors to overcome the problem of vegetation occluding tree trunks and thereby resulting in noisy 2D LiDAR scans on which fitting lines is challenging \cite{Bell2016,Zhang2013}. Since commercially available 3D LiDAR sensors are expensive, \cite{Zhang2013} used a planar LiDAR and added a rotational DoF to capture 3D point clouds but this approach adds complexity to the system. A push-broom based modification to 2D LiDAR was another approach that was tested to capture 3D point clouds \cite{Baldwin2012RoadPriors}. Row crops environments such as corn, sorghum are very different from orchards. Unlike orchards where individual trees are separated by a large distance and also subsequent rows have a large gap between them, canopy is dense in row crops and also the row spacing is limited to around 0.8~m in corn/sorghum. This results in significant occlusion and clutter present in the LiDAR scans making it challenging to find the crop rows in the scans. \cite{Hiremath2014} developed a 2D LiDAR based navigation system for corn using particle filter to handle noise present in the environment, but since it is an over-the-canopy platform it is limited to only very early growth stage of the crops. \cite{Troyer2016} used a 2D LiDAR to develop navigation system for corn but they did not validate their approach in real field/plants. In late growth stage corn and sorghum, as a robot navigates under the canopy, the above mentioned challenges of clutter and noise in LiDAR scans. This requires careful analysis of the scans to detect the crop rows as shown by \cite{higuti2019under} and also using sensor fusion with encoders and IMU can improve the robustness of under-canopy as shown in our results.

\section{System Design}

\begin{figure}[h]
    \centering
    \includegraphics[width=1.0\textwidth]{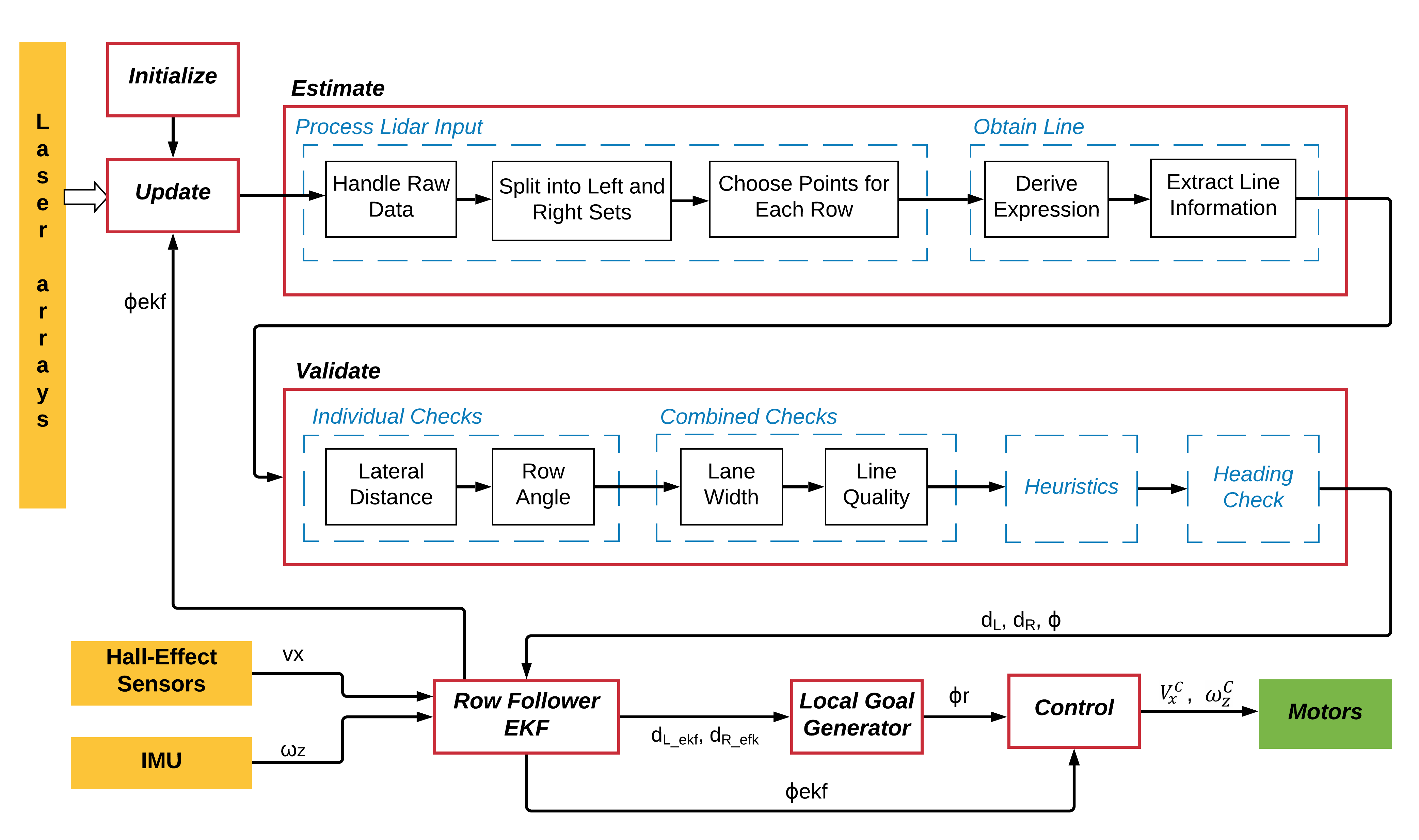}
    \caption{Navigation system diagram}
    \label{fig:block-diagram}
\end{figure}

The proposed navigation system (Fig. \ref{fig:block-diagram}) is composed by seven steps: \textbf{Initialize}, \update, \textbf{Estimate}, \textbf{Validate}, \ekfpl, \textbf{Local Goal Generator} and \control. The first four steps make up the perception subsystem that uses the raw LiDAR measurements from the surroundings to build two virtual representation of the rows. These virtual walls are used to estimate the distances ($d_R$ and $d_L$), and robot's heading ($\phi$). Then \ekfpl\ gets the linear ($v_x$) and angular ($\omega_z$) velocities estimations from the Hall-Effect Sensors and the Inertial Measurement Unit (IMU) to correct $d_R$, $d_L$ and $\phi$. Then the corrected distances are used by the \textbf{Local Goal Generator} to generate a reference heading ($\phi _r$). Finally, the \control\ calculates the control action ($\omega c_z$) using $\phi$ and $\phi _r$.

\subsection{Robotic platform}

The TerraSentia robot system is a compact and autonomous robot designed particularly for high-throughput phenotyping. It was conceived by Distributed Autonomous Systems Laboratory (DASLAB) at the University of Illinois at Urbana Champaign, and since 2017, it has been manufactured and commercialized by EarthSense Inc. TerraSentia has been designed to be rugged for all-season field deployment, yet remaining light-weight (16.55 Kg with battery) and low-cost. The robot dimensions are 0.54~m x 0.32~m x 0.35~m (length x width x height) and because of this, it can operate under the canopy in crops with row spacing as little as 0.4~m. Its body, which was design to ensure the system does not harm young plants, is made of hard plastic and reinforced by a lightweight metal frame.

The robot has four Maytech Brushless Outrunner Hub Motor with hall-effect sensor to drive each of its four wheels which are built through additive manufacturing with polylactic acid. Each motor is positioned in the middle of the wheel and driven by a custom version of the VESC4 motor controllers.  For navigation, Terrasentia has embedded a Bosch BNO055 IMU (Inertial Measurement Unit) and a Zed-F9P GPS (Global Positioned System) manufactured by Ublox. These sensors are integrated in a main board manufactured by Earthsense.   

Two 2D Hokuyo LiDAR (UST-10LX) scan the environment. The first one is positioned in the front part and it provides a surroundings' horizontal scan, which serves as input of the proposed navigation system. The second one is positioned in the rear part in order to provide a vertical scan that is used to make an offline 3D crop reconstruction. Additionally, four wide angle Full HD USB Camera Modules 1080P record videos to extract meaningful phenotype features using the automatically time- and geo- tagged data from them. This includes plant count, stem width, leaf area index under the canopy, plant height and disease detection \cite{KayacanGirish2018,Choudhuri2018CropEnvironment}.

All sensors are connected to a Raspberry Pi 3, which is the main computer of the robot. It acquires the readings from the front LiDAR, IMU and hall-effect sensors, runs the proposed navigation system, and generates the desired control signals (desired robot's angular and linear velocities). Then it transforms the control signals to the desired velocity for each motor and sends these values to the VESC4 controllers as PWM (Pulse Width Modulation) signals to drive the motors. Additionally, the robot has a Intel NUC i7 computer with 500GB SSD and 16GB RAM to store all collected data. 

\begin{figure}[ht]
\begin{subfigure}{.5\textwidth}
  \centering
  % include first image
  \includegraphics[width=.7\linewidth]{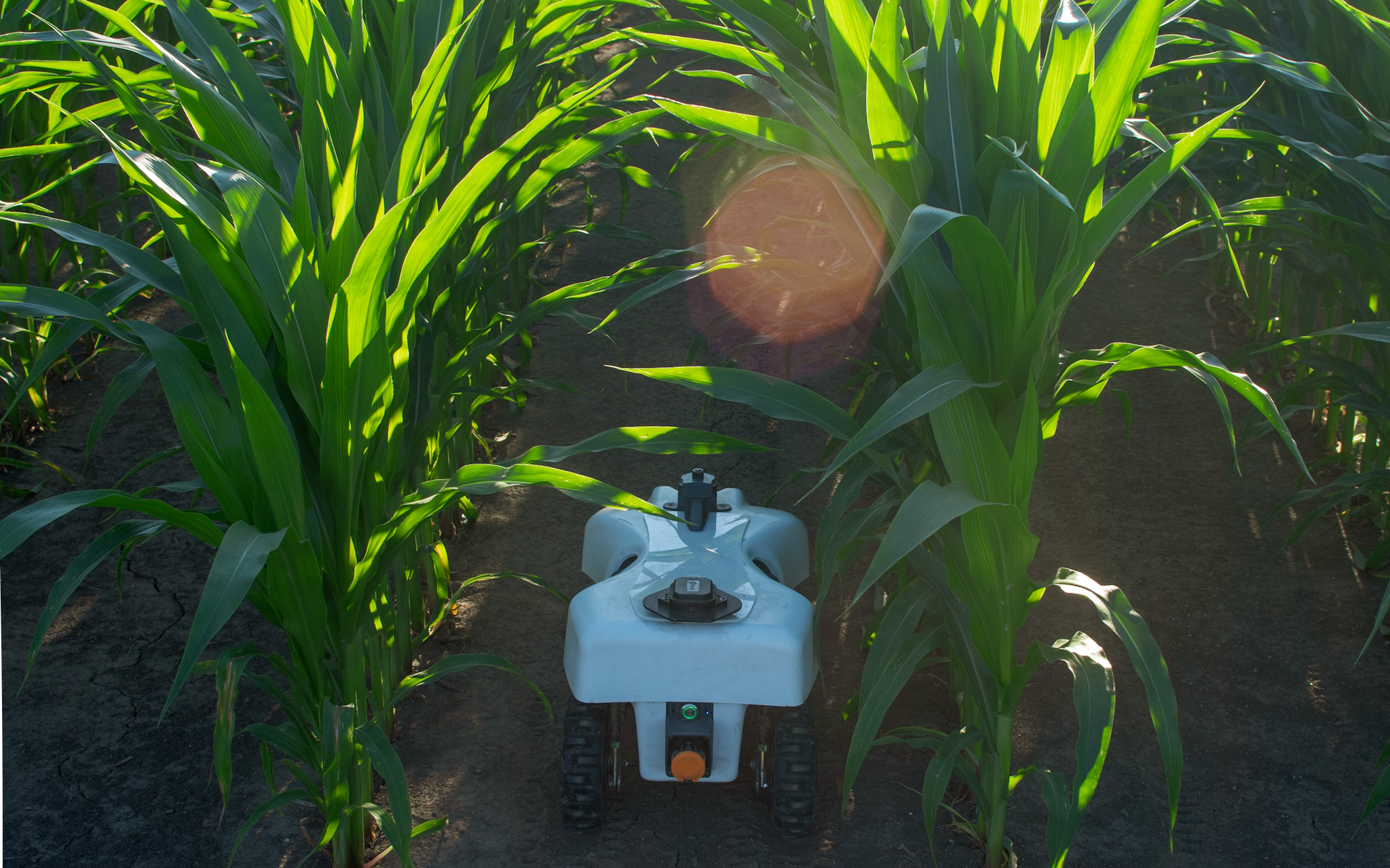}  
  \caption{}
  \label{fig:terra2020}
\end{subfigure}
\begin{subfigure}{.5\textwidth}
  \centering
  % include second image
  \includegraphics[width=.7\linewidth]{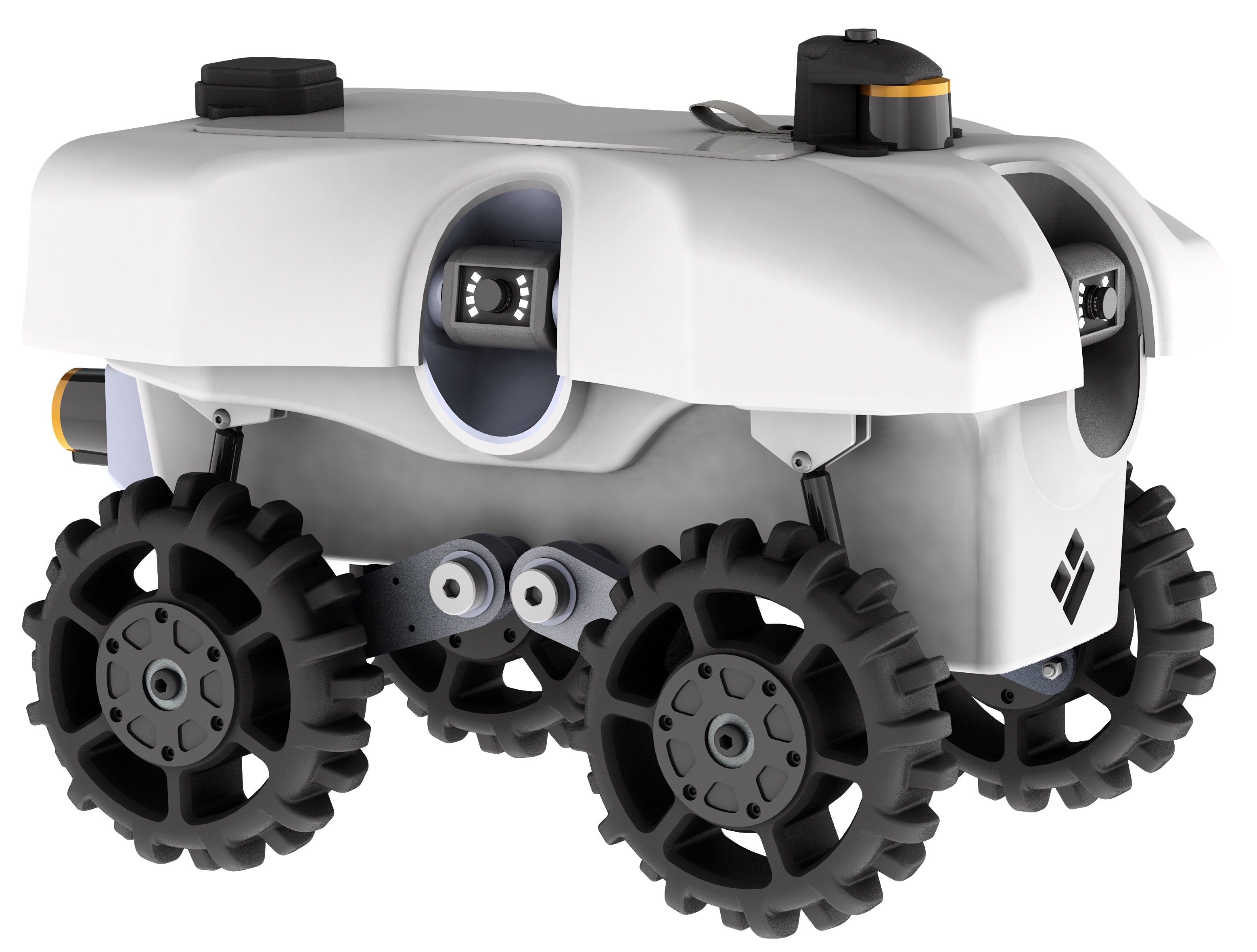}  
  \caption{}
  \label{fig:CAD2020}
\end{subfigure}
\caption{a-) TerraSentia robot in a corn crop, b-) CAD drawing of the Robot}
\label{fig:Terrasentiaphotocad}
\end{figure}

\subsection{Perception Subsystem}

The perception subsystem is composed by 4 stages: \textbf{Initialize}, \update, \textbf{Estimate} and \textbf{Validate}. In the first stage, the configuration files are parsed and stored as variables in the appropriate classes, and the robot’s embedded devices are initialized. These configuration files contains the information about the physical characteristics of the robot, threshold values, function limits, main crop characteristics and other initial conditions \cite{higuti2019under}. 

The next stage is the \update. In this stage, the LiDAR readings and the robot’s current orientation angle $\phi$ are obtained and sent to the \textbf{Estimate} stage. Also the update stage received the constants set from \textbf{Initialize} stage during the first loop. All these variables are used in the \textbf{Estimate} stage  in order to filter out extraneous information and creating a data set to determine the new values for $d_R$, $d_L$ and $\phi$. 

\begin{figure}[thpb]
    \centering
    \includegraphics[width=0.486\textwidth]{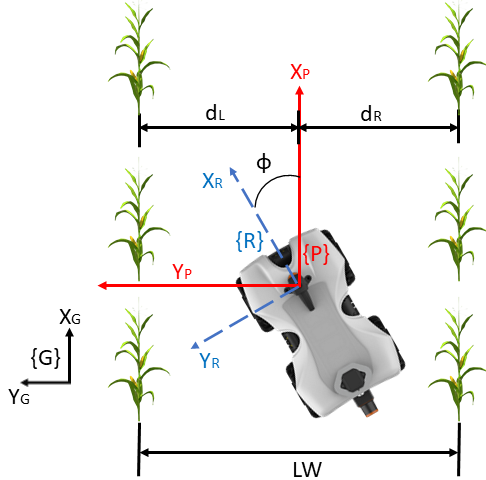}
    \caption{Global {G}, Path {P} and robot {R} reference frames. The origins of {P} and {R} are fixed at LiDAR geometric center. Robot y‐axis ($X_R$) is aligned with robot’s longitudinal axis, path y‐axis ($X_P$) is parallel to rows. $\phi$ is the angle between these two x‐axis. $d_R$ and $d_L$ are the orthogonal distances between the center of sensor and each adjacent row.}
    \label{fig:frames}
\end{figure}

In the first step ($Process LiDAR Input$) of the \textbf{Estimate}, the raw LiDAR readings are projected from Polar to Cartesian coordinates in the robot Local frame (\{R\} in Fig. \ref{fig:frames}). Then they are transformed from Local to the Path Frame (\{P\} in Fig. \ref{fig:frames}) using  robot's heading ($\phi$) to compensate the robot’s orientation deviations. \{P\} is a translation of the Global coordinate frame \{G\}, which is located at a corner of the corn crop, to the origin of \{R\}. \{P\} and \{R\} have their origins in the geometric center of LiDAR, the first has its x‐axis ($X_P$) always parallel to one of the rows and the latter rotates with the robot. Also $\phi$ is considered as the angular difference from $X_P$ to $X_R$.

As a single LiDAR scan has 1081 distances on a plane with $270\degree$ angular range, then there are some readings without any relation with the rows. For that reason, the readings inside of a fixed rectangle $(y \in (-D_y , D_y ) , x \in ( D_x0 , D_x1 ))$ around the origin of the frames \{R\} and \{P\} are used in the next steps. $L_x$ is defined to consider the points related to neighboring rows while the points of the farther rows are discarding. $D_x0$ was defined as 0.1m in order to discard the points behind of the $Y_P$ and $D_x1$ was 1.9m. Opposed to previous work, the distance readings behind LiDAR are not considered anymore because the leaves located in the rear part of the robot usually form a dense curved shape which allows multiple starting points for straight lines generating wrong estimations of the distances and robot's heading. We observed that this problems is more common when the robot is used to navigate in denser crops. Once the LiDAR readings are filtered, they are split in two sets: right and left side. The use of $\phi$ to rotate the raw readings helps to guarantee that one of the captured rows is almost parallel to $X_P$ axis. This condition allows the use of an histogram to detect where the relevant information about the position of each row is. The histogram generates two peaks (one for the right side and the other one for the left side) and all the readings around these peaks are considered to obtain the virtual walls in the $Obtain Line$ step. 

In the $Obtain Line$ step, the lest minimum squares linear regression is used to obtain the virtual lines. Also the orthogonal distance to such segments provide the values of $d_R$ and $d_L$. Some properties (such as length, standard deviation, difference between current orthogonal distance/slope with previous ones) of the virtual lines are used to determine the quality of the estimation, and they are used by a heuristic to define if the distances are valid or not. If they are valid then their values are sent to the \ekfpl\ but if they are not valid then their last valid value are used in the EKF. The validation process happens in the \textbf{Validate} stage.

% Include EKF part
\subsection{Extended Kalman Filter (EKF) for Perception}

To improve performance of lateral distances and heading estimations, this paper takes advantage of the complementary characteristics of Perception algorithm that estimates the absolute position of the robot in relation to the crop row, and robot's embedded sensors that estimate relative robot's displacement at every time instant. Thus, an Extended Kalman Filter (EKF) was chosen to provide a better estimation of the true values given a fusion between available sensors.

To provide a reference frame for the mechanization equations that represent the movement of the robot in relation to the rows, a new coordinate frame is defined. This new coordinate frame $\{N\}$ assumes two straight rows in form of an aisle with constant distance between the rows, as shown in Figure \ref{fig:frames}. The inertial coordinate frame $\{N\}$ is defined with origin localized in the entrance of the corridor. The x-axis of coordinate frame $\{N\}$ indicates the longitudinal distance along the path. The y-axis indicates the perpendicular distance across this same path and indicates the robot's cross-track error along the path. For reference purposes, the z-axis points upwards.

A differential system of equations can be written to represent the movement of the robot along this path, as shown in Equation \eqref{eq:environment}. $p$ is
the three-dimensional position of the robot in relation to the origin of $\{N\}$, $C_{\{R\}}^{\{N\}}$ is the rotation that aligns the frame $\{R\}$ to the frame $\{P\}$, and $v$ is the three-dimensional robot's velocity written in terms of the local frame  $\{R\}$.

\begin{equation} \label{eq:environment}
    \dot{p} =  C_{\{R\}}^{\{N\}} v
\end{equation}

We assume displacements along inertial z-axis and rotations around x and y-axes are negligible, so a planar version with only the x and y-axes is used. Equation \eqref{eq:rotation} shows the rotation matrix that transforms from local robot's body frame to the $\{N\}$ inertial frame using this planar assumption.

\begin{equation} \label{eq:rotation}
     C_{\{R\}}^{\{N\}} = 
    \begin{bmatrix}
        cos \phi & sin \phi\\
        -sin \phi & cos \phi
    \end{bmatrix}
\end{equation}

And then 

\begin{equation} \label{eq:mechanization}
    \begin{bmatrix}
        \dot{p_x} \\
        \dot{p_y}
    \end{bmatrix}
    = 
    \begin{bmatrix}
        v_x cos \phi + v_y sin \phi\\
        - v_x sin \phi + v_y cos \phi
    \end{bmatrix}
\end{equation}

$p_y$ is a metric of cross-track error, and for this specific problem, we define cross-track error as $0.5(d_R-d_L)$. Also, the sum $d_R + d_L$ is assumed constant and it's equal to the crop lane width $LW$. Since the lateral distances $d_R$ and $d_L$ are the values of interest in this project, the differential equation can be rearranged to highlight these metrics, as shown in Equation \eqref{eq:new-mech}.

\begin{equation} \label{eq:new-mech}
    \begin{bmatrix}
        \dot{p_x} \\
        \dot{d_L} \\
        \dot{d_R}
    \end{bmatrix}
    = 
    \begin{bmatrix}
        v_x sin \phi + v_y cos \phi \\
        - v_x sin \phi + v_y cos \phi \\
        v_x sin \phi + v_y cos \phi
    \end{bmatrix}
\end{equation}

Further extension of state estimation can be given by adding the estimation of the angle $\theta$, where $\dot{\theta} = \omega$. Furthermore, the state $p_x$ is not of interest in this project, so it will not be considered in this formulation. In addition, because a non-holonomic model formulation is adopted to represent robot's kinematic, the lateral velocity $v_y$ is also disregarded. The control variables $v_x$ and $\omega$ are chosen as inputs of our differential system, such that $u = [v_x, \omega]^T$. 

\begin{equation} \label{eq:final-mech}
    \begin{bmatrix}
        \dot{d_L} \\
        \dot{d_R} \\
        \dot{\phi}
    \end{bmatrix}
    = 
    \begin{bmatrix}
        - v_x sin \phi \\
        v_x sin \phi \\
        \omega
    \end{bmatrix}
\end{equation}

Using the differential system stated at Equation \ref{eq:final-mech} and solving it numerically using forward Euler method, a solution model is encountered, where $x_{k} = f(x_{k-1}, u_{k-1})$. The input $u_k$ is provided by embedded sensor measurements, which we define as $u_k = (v_x, \omega)$. Because the model is updated using noisy sensor measurements, and these measurements are susceptible to uncertainties, then we assume a process uncertainty term $\Delta_k$ added to the prediction model. The resultant equation is $x_{k} = f(x_{k-1}, u_{k-1}) + \Delta_k$, and the uncertainty is assumed as a zero mean Gaussian noise. This solution model is used to predict future robot positions according to the created coordinate system $\{N\}$.

\begin{equation}
    \begin{array}{ccl}
        x_k & = & f(x_{k-1}, u_k) + \omega_k \\
        z_k & = & h(x_k) + \nu_k 
    \end{array}
    \label{eq:ekf}
\end{equation}

To correct state prediction from the obtained mechanization model, an Extended Kalman Filter framework was chosen to account for the uncertainties in distance and heading estimation \cite{Thrun2002}. We can model the process with equation \eqref{eq:ekf} where actual state $x_k$ is a function $f(\cdot)$ of past state $x_{k-1}$ and control inputs $u_k$, as shown in equation \eqref{eq:stateFunction}. $dt$ is the time step using in the forward Euler method and represents the time taken to run one EKF iteration. 

\begin{equation}
    f(x_{k-1}, u_k) = 
    \begin{bmatrix}
        d_{L_{k}} \\
        d_{R_{k}} \\
        \phi_{k}
    \end{bmatrix}
    =
    \begin{bmatrix}
        d_{L_{k-1}} - v_{x} \sin\phi_{k-1} dt \\
        d_{R_{k-1}} + v_{x} \sin\phi_{k-1} dt \\
        \phi_{k-1} + \omega dt
    \end{bmatrix}
    \label{eq:stateFunction}
\end{equation}

In order to calculate the predicted covariance estimate, the Jacobian of function $f(\cdot)$ evaluated at each time step is used. The Jacobian matrix is shown in equation \eqref{eq:Jf}. The function $h(\cdot)$ relates the state $x_k$ to the measurement $z_k$, and they are directly obtained from the LiDAR-based perception model as shown in equation \eqref{eq:hFunction}.The Jacobian if function $h(x_k)$ to calculate the updated covariance estimate is the identity matrix. In our implementation, we assume the process and measurement covariance matrices are constant, and therefore $\mathbf{Q_k} = \mathbf{Q}$ and $\mathbf{R_k} = \mathbf{R}$.

\begin{equation}
    F_k = 
    \begin{bmatrix}
        1 & 0 & -v_{x} \cos\phi_{k} dt \\
        0 & 1 & v_{x} \cos\phi_{k} dt \\
        0 & 0 & 1
    \end{bmatrix}
    \label{eq:Jf}
\end{equation}

\begin{equation}
    h(x_{k}) = 
    \begin{bmatrix}
        d_{L_{k}} \\
        d_{R_{k}} \\
        \phi_{k}
    \end{bmatrix}
    \label{eq:hFunction}
\end{equation}

In the prediction step, $f(\cdot)$ (Eq. \ref{eq:stateFunction}) estimates the state using the linear velocity $v_x$ and angular velocity $\omega _z$, respectively from wheel encoder measurements and embedded gyroscope sensor. Subsequently, in the update step, the innovation occurs by taking into account the calculated values of lateral distances and robot's heading that are obtained from the LiDAR point-cloud (either raw 2-D or processed 3-D).

\subsection{Local Goal Generator for Row Follower and Controller}

The Local Goal Generator is a function that simulates a vector field where the value only varies in x. The behavior of the field is designed to point the robot towards the middle of the lane and gradually turn the robot to the normal orientation as it approaches the middle. This reduces the oscillatory behavior that arises when the control action is only based on the position, as we can  easily correct oversteering. The vector field (Eq. \ref{eq:sf}) was defined as the arctangent of the derivative of Eq. \ref{eq:sf1} which is used to establish the trajectory that the robot should follow when it is in row.

\begin{equation}
    f(x) ={\left(\frac{b}{\left|\frac{e_d}{e}\right|}\right)^c}
    \label{eq:sf1}
\end{equation}

\begin{equation}
    \phi_r= \arctan{\left(\frac{df(x)}{dx}\right)} = -\arctan{\left(c\left(\frac{e_d}{e^2}\right)b^c\left(\frac{1}{\left|\frac{e_d}{e}\right|}\right)^{c+2}\right)}
    \label{eq:sf}
\end{equation}

\begin{figure}[ht]
\begin{subfigure}{.48\textwidth}
  \centering
  % include first image
  \includegraphics[width=.8\linewidth]{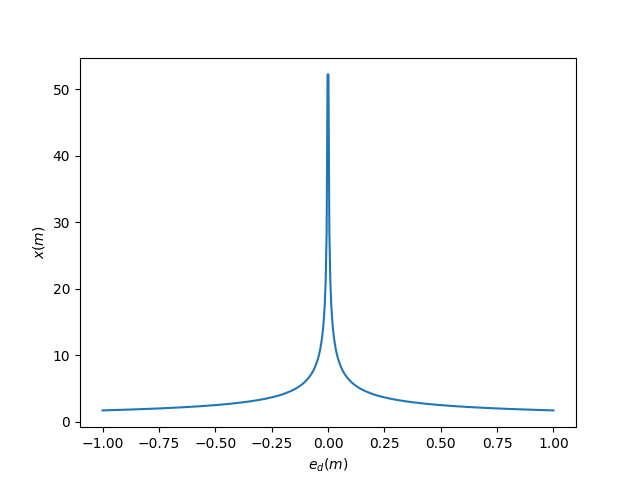}  
  \caption{}
  \label{fig:trajectory}
\end{subfigure}
\begin{subfigure}{.48\textwidth}
  \centering
  % include second image
  \includegraphics[width=.8\linewidth]{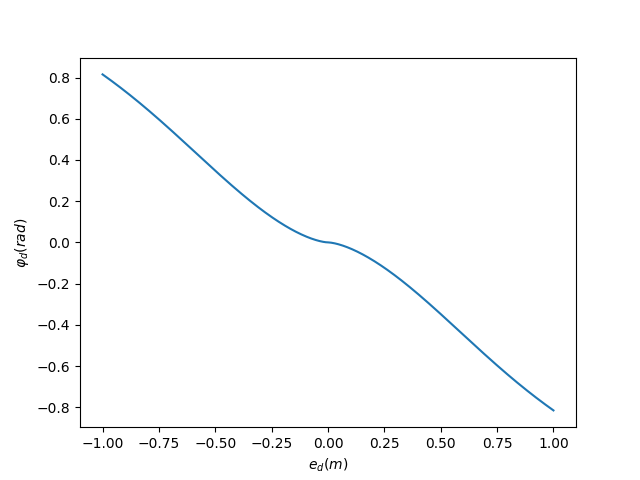}  
  \caption{}
  \label{fig:heading}
\end{subfigure}
\caption{a-) Desired trajectory given by Eq. \ref{eq:sf1}, b-) Desired behaviour of $\phi_r$}
\label{fig:potentialfield}
\end{figure}

Where \textbf{e} is the curve width constant, \textbf{a} is the curve smoothness constant, \textbf{c} is the curve deadband constant, $d_{ref}$ is the reference distance and $e_d$ is the error distance which is defined as $e_d = d_{ref} - 0.5(d_R$\underline{\space}$_{ekf}$ $- d_L$\underline{\space}$_{ekf})$. $d_{ref}$ is zero because it is expected that robot travels following the lane in order to minimize possible collisions with lateral rows. Fig. \ref{fig:trajectory} shows the trajectory given by Eq. \ref{eq:sf1} and Fig. \ref{fig:heading} shows the desired behavior of $\phi_r$.

The output of the Local Goal Generator is the target heading ($\phi_r$). The next step of the navigation system is a PID controller used to keep the robot in the path center. Its outputs are $\phi_r$ and$\phi_ekf$ and its output is an angular velocity ($\omega_z^C$) command which is sent, together with the linear speed ($V_x^C$) command, to the motors.

\section{Success Metric}

% \miss{we should think about placing visual tags to be captured by lateral cameras and using them as ground truth}
A successful autonomous run of mobile robots is often subjected to assessment of how well it followed the desired path. \cite{Higuti2018UnderNavigation} discusses two derived values from perception subsystem lateral distance outputs: Cross track error and Lane width. The first is the distance between robot and the reference path, i.e. the middle of the lanes. It is given by half of difference between right and left lateral distances, $CTE = 0.5(d_R-d_L)$. The second is the measured navigable space and it can be compared to nominal row spacing when seeds were planted. It is given by the sum of lateral distances $Lw = d_R + d_L$.

Such values raise the need of ground truth, which will either give the exact robot's positioning over time to compare with cross track error or provide what lane width should be. Nevertheless, ground truth encounters several obstacles in crop environments. Commonly used for outdoor applications, RTK-GNSS suffers from signal degradation due to loss of satellite view and multi-path under canopy. A stick to raise GNSS antenna above coverage is not feasible since plants have more than 2~m and such stick would get entangled and heavily oscillate due to ground unevenness. Markers placed on ground would require constant maintenance because the robot would go over it, which also raises a ground condition change. Since LiDAR itself provide raw distance measurements, a manual ground truth can be obtained from them and the results may be directly compared. For such task, one every 10 scans is analyzed and the visually best fitting line is annotated for each side, if available. 

Given that sensor operates on 40~Hz scan rate, manual ground truth is hardly scalable as field experiments increases in quantity and duration. Since robot is still in a phase where it is followed by an operator, the current autonomous mode is expected to relieve people from the tedious task of driving the robot while going through crop lanes. Therefore, the success metric adopted in this work is the \textbf{distance per intervention}: How much crop was covered between interventions, which is defined as situations where failures in perception or control subsystems required operator to recover manual control of the robot. It also counts situations where system fail soon after starting (bad start) and also failures to presence of gap, an important feature in research fields.

\section{Experimental Results}

The field experiments evaluated the perception and control subsystems on corn crops and they can be divided into two categories: 1-) Controlled Tests, and 2-) Uncontrolled Tests. 

The \textbf{Controlled Tests} were designed to compare performance of previous perception subsystem \cite{Higuti2018UnderNavigation}, further called PL, with the one proposed here, referred as PL+EKF. Besides comparison, another goal is to pinpoint the failure sources. Such experiments expose TerraSentia's autonomous navigation to diverse field situations as opposed to previous study, whose corn experiments were limited to well behaved lanes and sorghum experiments, although varied from well behaved to intensely cluttered, they were limited to three meters of lane length. 

The \textbf{Uncontrolled Tests} uses only PL+EKF and they have been performed by third parties with no specific knowledge about the perception and control subsystems. For such cases, the autonomous navigation is an auxiliary tool for another research (mainly collection of visual data along rows) in two USA states (Illinois and New York). These tests brought diversity of field and also different user cases as going through lane from begin to end could not be the main goal, but rather running few meters to capture a specific part of the crop for offline analysis.

\subsection{Experimental setup}

For the validation tests (Controlled and Uncontrolled tests) presented in this paper, the core Perception Subsystem ran with same configurations as the one presented in \cite{Higuti2018UnderNavigation} except for nominal lane width, which was set to 0.75~m since all fields share this characteristic within 0.10~m interval, and desired forward speed, which raised from 0.22~m/s to 0.7~m/s. 

The EKF model and measurement error covariance matrices, $\mathbf{Q_k}$ and $\mathbf{R_k}$ respectively, were set as constant diagonal matrices whose values were empirically determined on field tests. $\mathbf{Q_k}$ was set to $diag\left(\begin{smallmatrix} 0.001 & 0.001 & 0.01 & 0.01 \end{smallmatrix}\right)$ while $\mathbf{R_k}$ was set to $diag\left(\begin{smallmatrix} 0.05 & 0.05 & 0.5 & 0.5 \end{smallmatrix}\right)$. The parameters of the Local Goal Generator were determined in the same way than the matrices  $\mathbf{Q_k}$ and $\mathbf{R_k}$ and their values are $b=3.8$, $c=0.55$ and $e=0.7$.

\subsection{\textbf{Controlled Tests}}

\begin{figure}[ht]
\begin{subfigure}{.5\textwidth}
  \centering
  % include first image
  \includegraphics[width=0.85\linewidth]{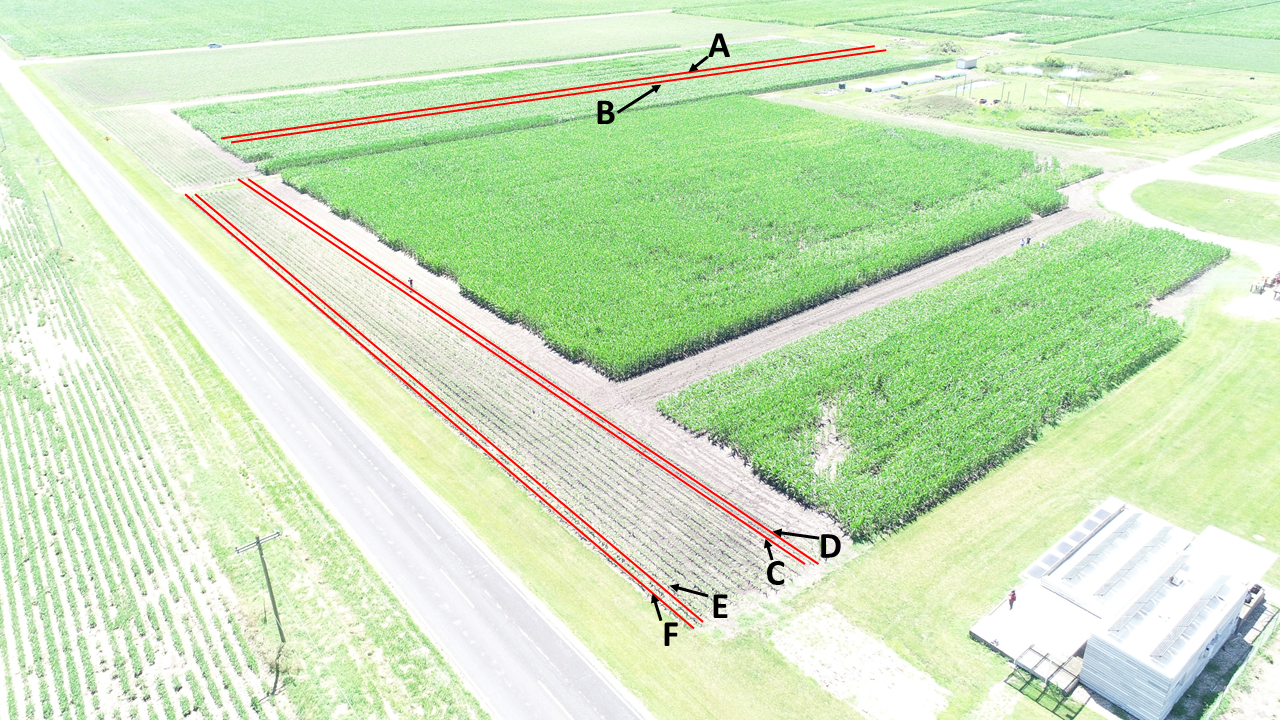}
  \caption{}
  \label{fig:AutoFarm}
\end{subfigure}
\begin{subfigure}{.5\textwidth}
  \centering
  % include second image
  \includegraphics[width=1.0\linewidth]{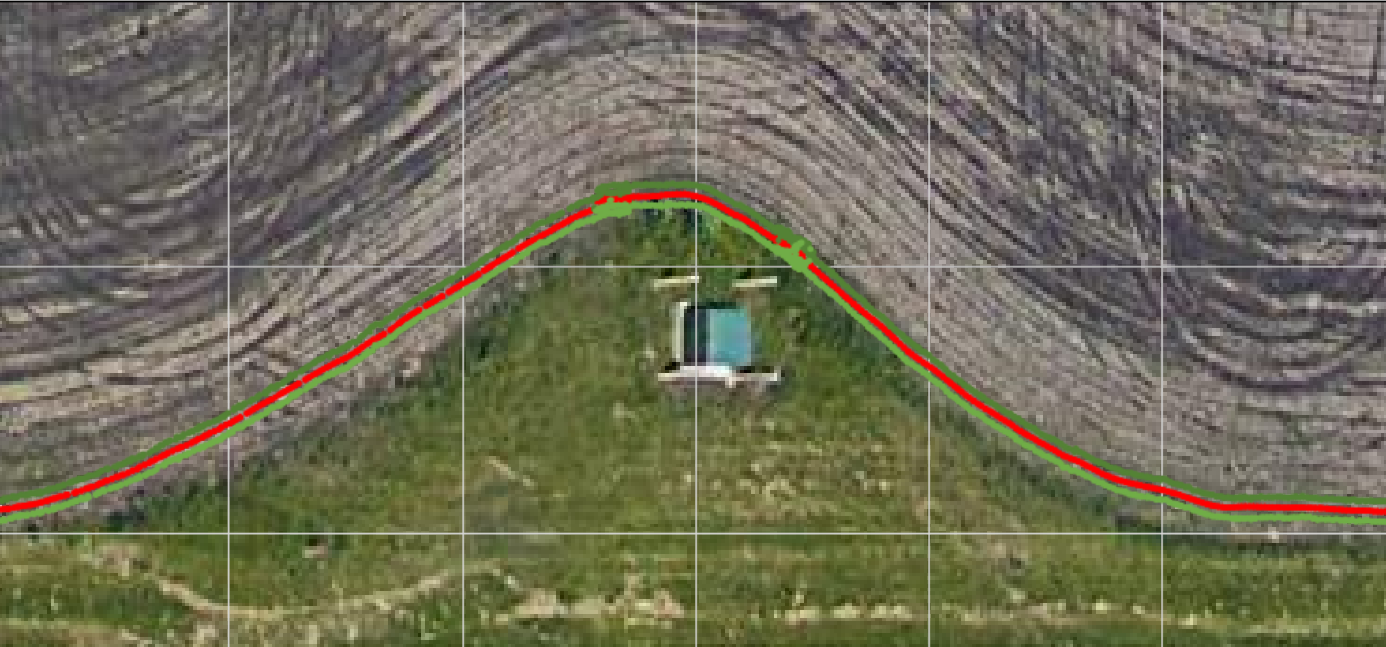}  
  \caption{}
  \label{fig:productionFarm}
\end{subfigure}
\caption{a-) UIUC Illinois Autonomous Farm research field (\autofarm); b-) Production Field farm with curve. The red lines indicates the rows used for the \textbf{Controlled Tests}}
\label{fig:Farms}
\end{figure}

The \textbf{Controlled Tests} were performed in two fields (Figure \ref{fig:Farms}): 1-) UIUC Illinois Autonomous Farm research field (further referred as \autofarm), and 2-) Production field farm with curve. Both fields  are located in Urbana-Champaign, Illinois, USA, and their mains characteristics are: 1-) Nominal row spacing equal to 0.75~m, 2-) Plant spacing around 0.1~m except for lanes with random sparseness or artificially induced gaps and 3-) Growth stage R4-R5. 

\subsubsection{\textbf{Controlled Tests} in \autofarm}

\begin{figure}[ht]
\centering
  \begin{subfigure} {0.49\linewidth}
    \centering
    \caption{}
    \includegraphics[width=50mm]{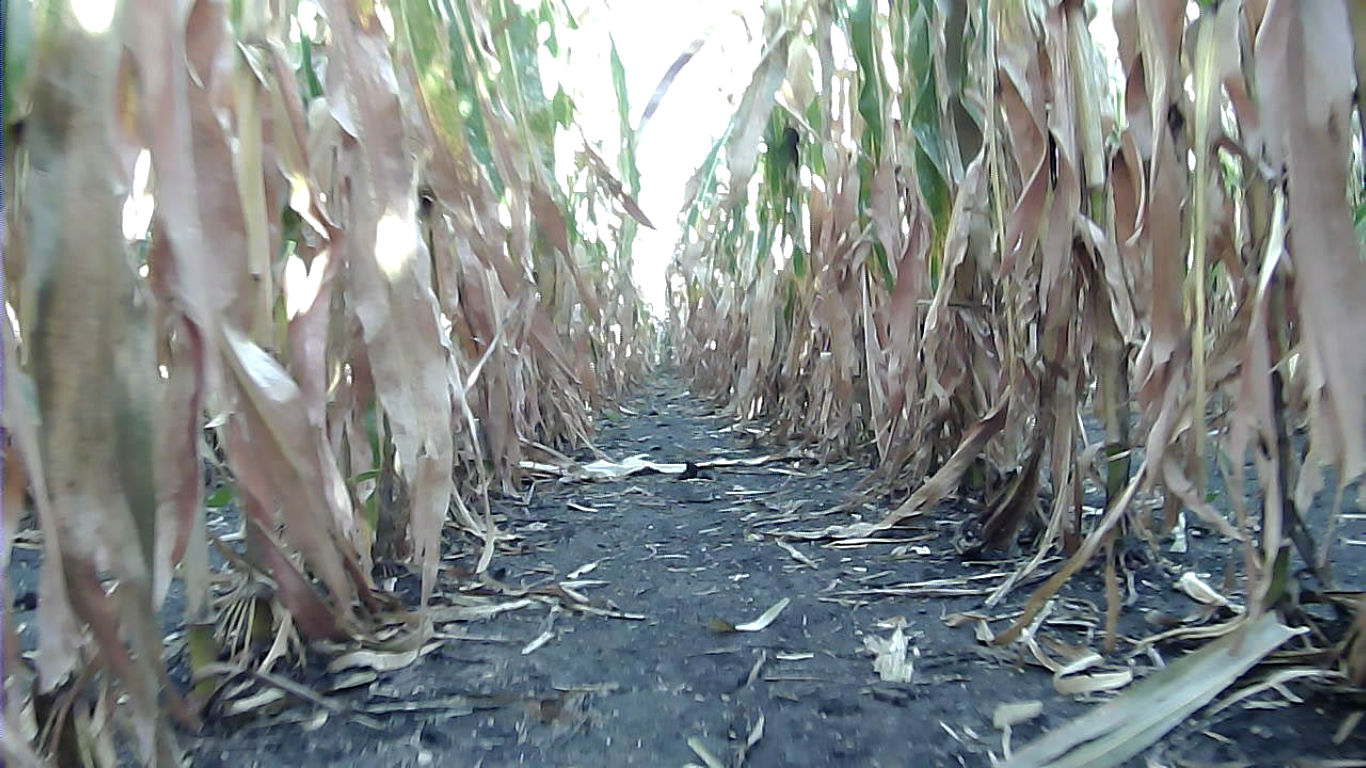} 
  \end{subfigure}
  \begin{subfigure} {0.49\linewidth}
    \centering
    \caption{}
    \includegraphics[width=50mm]{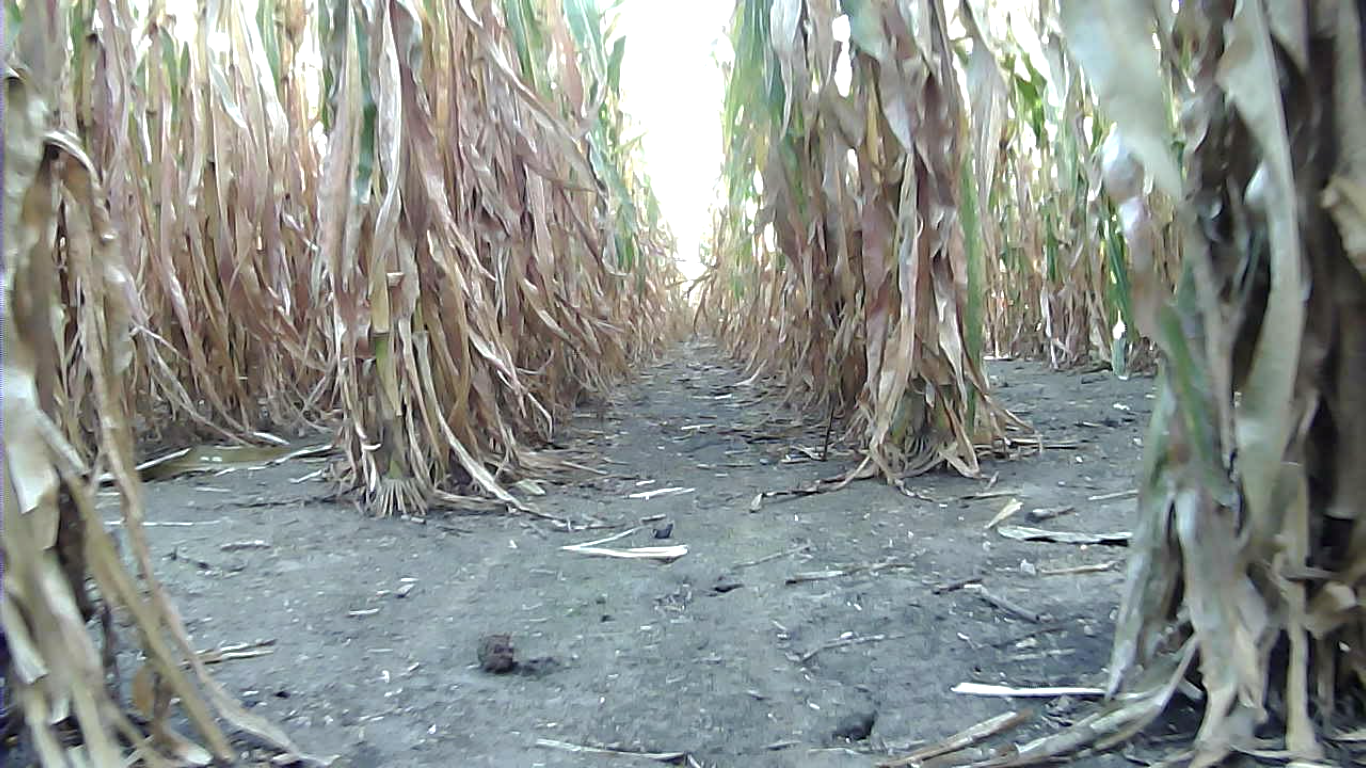} 
  \end{subfigure}
  \begin{subfigure} {0.49\linewidth}
    \centering
    \caption{}
    \includegraphics[width=50mm]{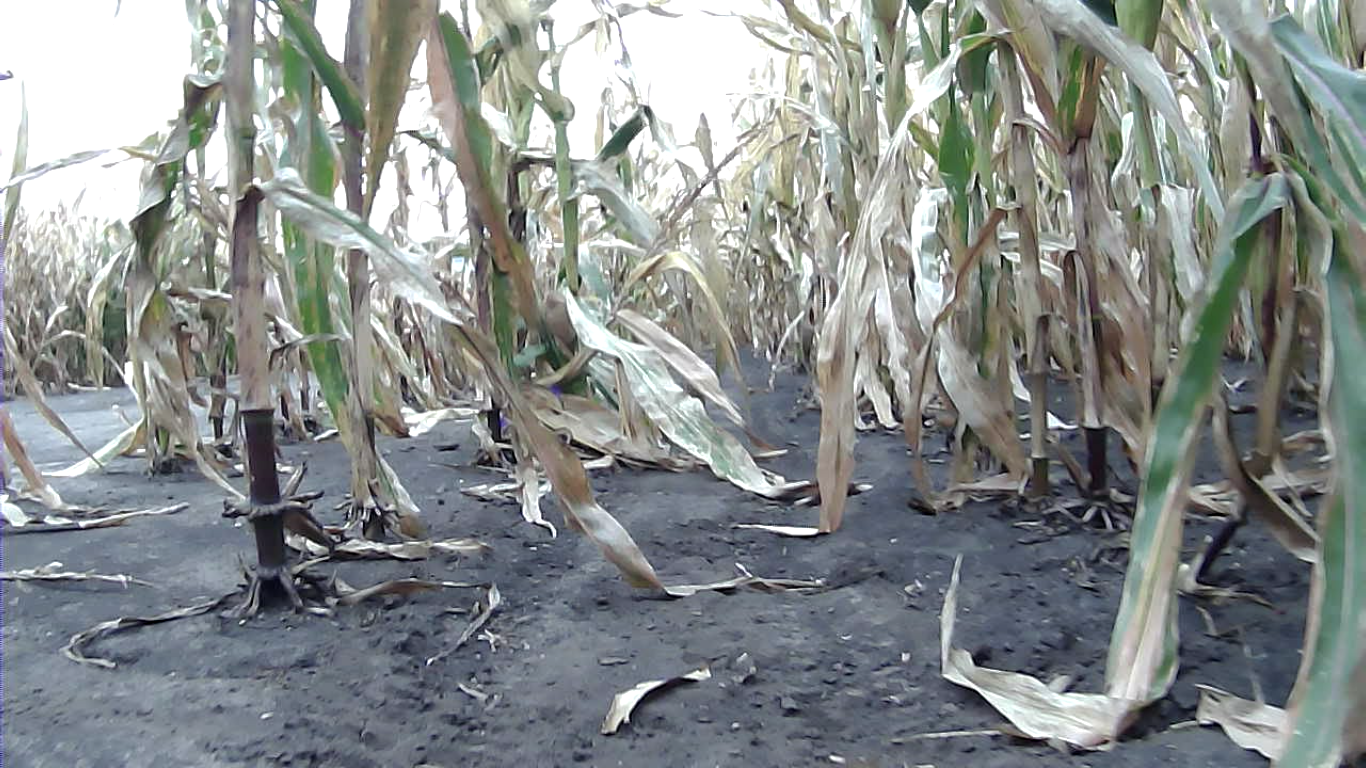}
  \end{subfigure}
  \begin{subfigure} {0.49\linewidth}
    \centering
    \caption{}
    \includegraphics[width=50mm]{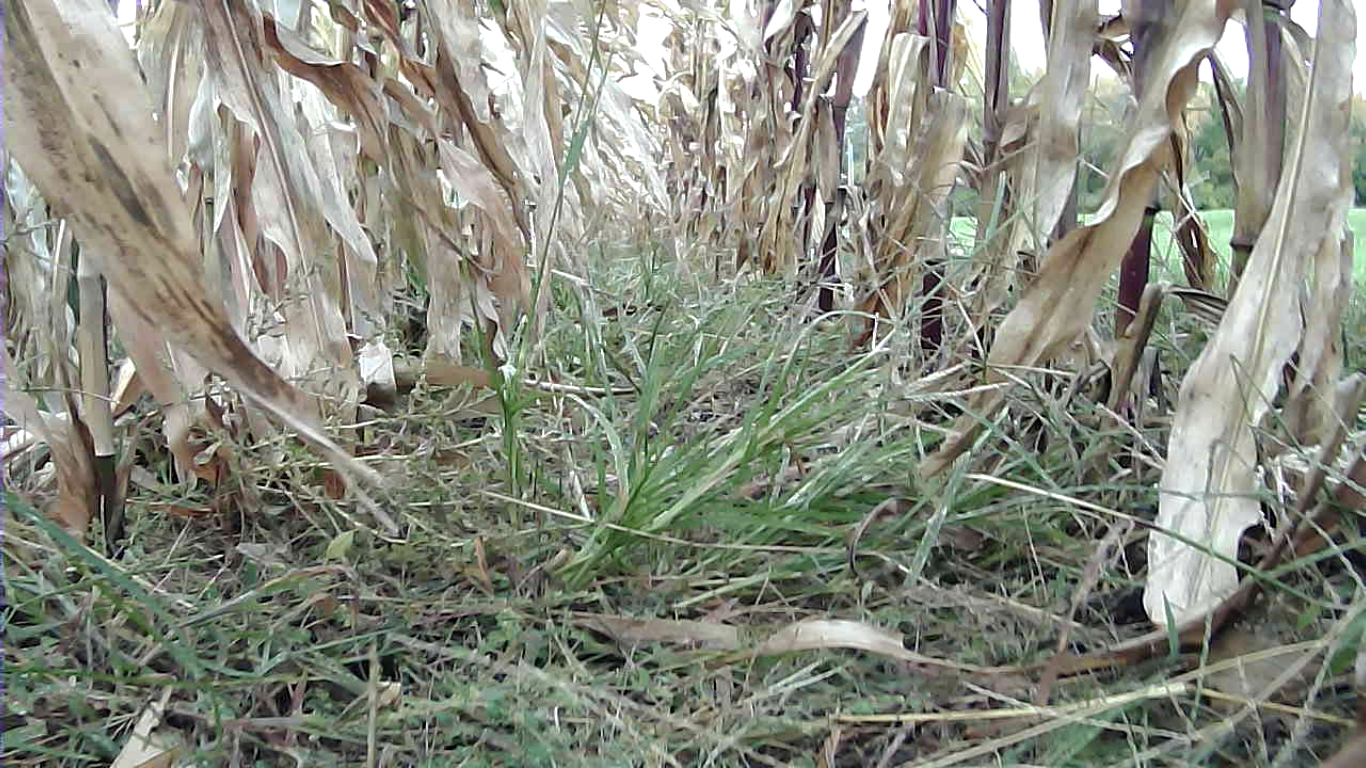}
  \end{subfigure}
\caption{Field tests occurred in Autonomous Farm, which contains a corn crop with (a) Both rows continuous from begin to end; (b) Artificially introduced gaps (c) Sparse rows with missing plants and fallen stalks (d) Less cared lanes with grass, weeds and clutter.}
\label{fig:scenarios}
\end{figure}

Five situations were evaluated in \autofarm. The first scenario (lane A) is the best case where both rows exist from begin to end of lane and they are straight and mostly parallel (Fig. \ref{fig:scenarios}a). The second (lane B) is similar to first but it has 12 gaps from 0.5~m to 1~m and 1 gap of 2~m (Fig. \ref{fig:scenarios}b). The length of both rows were 220~m. Third scenario used the same lanes and the LiDAR readings were acquired with different update rates of 10~Hz and 5~Hz. Finally, fourth and fifth were 125~m length rows characterized by sparseness and fallen stalks (Fig. \ref{fig:scenarios}c) or presence of high grass and weeds (Fig. \ref{fig:scenarios}d). 

Lane $A$ provides a baseline as it is the less externally influenced. Table \ref{tab:comparison_gap} shows that five interventions happened in runs 1 and 3. Except the second one for Perception LiDAR without EKF (PL), all of them were end-of-lane interventions, i.e. robot got to the end of the lane and required manual operation to go to next lane. That intervention was a collision after heading estimation diverged due to most of both sides being occluded by leaves in the middle of the path.

\begin{figure}[ht]
    \centering
    \includegraphics[width=.98\linewidth]{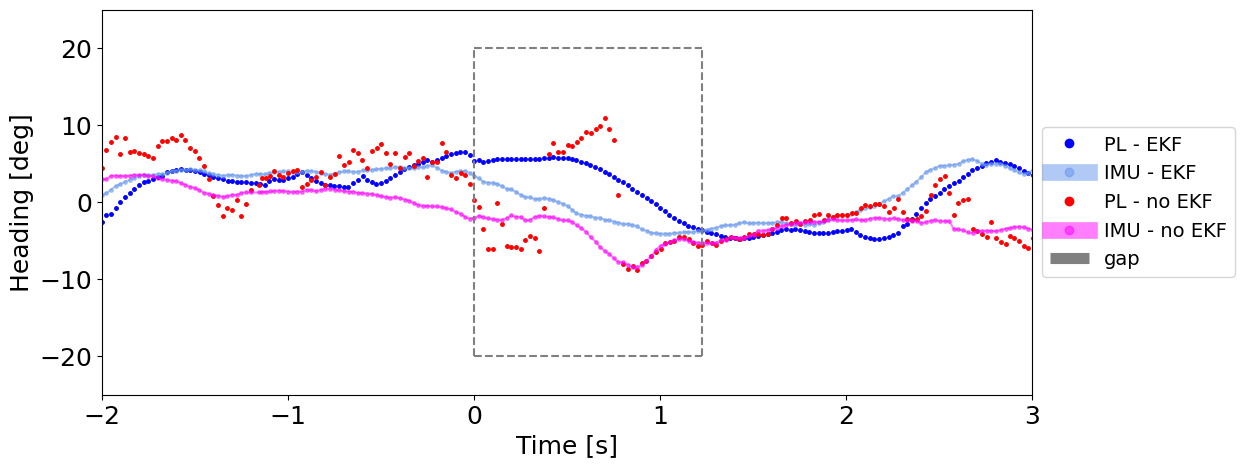}
    \caption{Heading analysis around gap (gray dashed box) for both runs in lane 2(B).}
    \label{fig:fig_heading_in_gap_ekf_vs_nonekf}
\end{figure}

\begin{table}[!h]
    \centering
    \caption{Number of interventions on lanes A (straight with gaps) and B (straight and continuous) in \autofarm.}
    \begin{tabular}{ p{1.2cm} p{0.4cm} p{0.4cm} p{0.4cm} p{0.4cm}}
       \cline{2-5}
        &\multicolumn{4}{ c }{Run (Lane)}\\
        \cline{2-5}
        &1(A) &2(B) &3(A) &4(B)\\
        \hline
        PL+EKF   &1 &1 &1 &2\\
        PL &1 &5 &2 &4\\
        \hline
    \end{tabular}
    \label{tab:comparison_gap}

\end{table}
All not end-of-lane interventions (8 of 12 for B) happened while crossing or soon after robot crossed the gap. This happened because of heading estimation error. In this case, as robot approaches a gap, there are fewer readings from the rows and they may not be from stems. These two factors combined destabilize heading estimation near gap. To illustrate this, Fig. \ref{fig:fig_heading_in_gap_ekf_vs_nonekf} shows heading estimates from Perception Subsystem (blue and red dots) and yaw measurements from IMU (light blue and magenta lines) for the same gap in both 1(B) experiments.

The gray dashed box represents the time window when robot crossed the gap. It may be noted that estimates worsen as robot approaches the gaps. Because of such bad heading, even though rows may be visible after gap, they are split into wrong sets and robot may be following bad fitted lines. Indeed, the estimates greatly differed from expected inside gap window but they recovered as robot approached the rows (right side of the dashed box in Fig. \ref{fig:fig_heading_in_gap_ekf_vs_nonekf}). Depending on how misaligned the robot left rows before gap, it may run diagonally, and can hit one of the rows, enter the neighbor lane or recover to the correct lane. This last option, when robot was already going away, is usually accompanied by a strong control action that causes an oscillatory driving and robot may crash soon after it enters the lane. Finally, Fig. \ref{fig:fig_heading_in_gap_ekf_vs_nonekf} also exposes the difference between Perception Subsystem with (blue) and without EKF (red). The estimates with EKF do not disperse and resemble the IMU readings. Even though differences are bigger around gaps, they are not as drastic as without EKF and they even allowed robot to cross all minor gaps. The only collision occurred in the 2~m gap, when robot lost track of the lane it was going and went away.

\begin{figure}[ht]
    \centering
    \includegraphics[width=.7\linewidth]{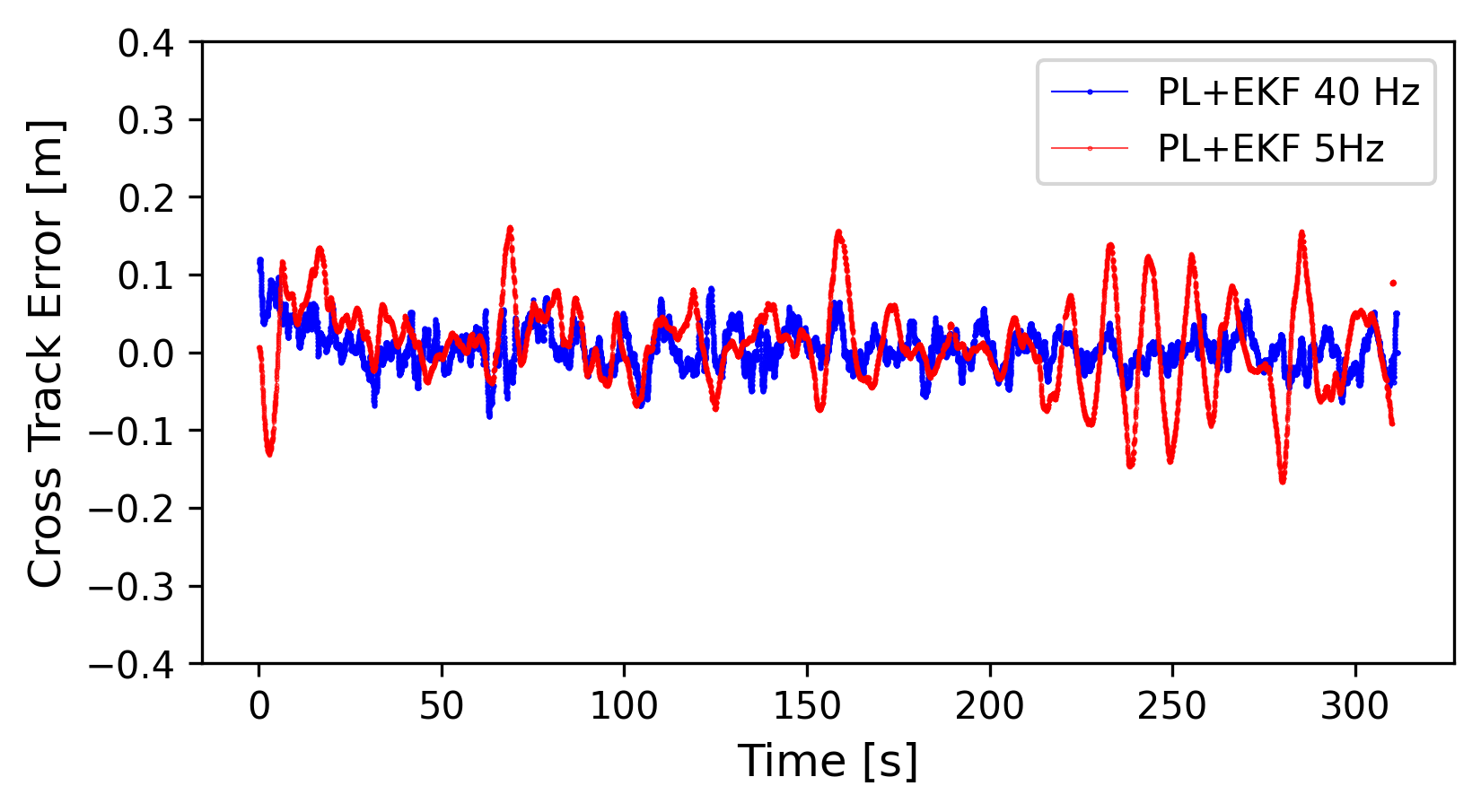}
    \caption{The cross track error comparison between PL+EKF with 5Hz (blue) and 40Hz (red).}
    \label{fig:cte_5Hz_ekf_vs_40Hz_ekf}
\end{figure}

Table \ref{tab:variable_lidar} shows the number of interventions in runs with variable LiDAR update rate. Since nominal update rate is 40~Hz, this frequency experiments agree with Tab. \ref{tab:comparison_gap}. It can be noted that number of interventions do not change when EKF is used. In lane B, both 5~Hz and 10~Hz presented three more interventions, all collisions, than 40~Hz. Besides the gap problem, less updated measurements prejudiced stabilization after occlusions because while a scan without occlusion is yet to be processed by Perception Subsystem, the robot blindly drives forward. Indeed, since calculation of control action requires updated estimates, while these are not available, the control remains the same.  For such reason coupled to the fact robot is traveling with 0.7~m/s, robot may overshoot the reference - middle of lane. This oscillating behaviour is clear on Fig. \ref{fig:cte_5Hz_ekf_vs_40Hz_ekf} between 200 and 300~s.

\begin{table}[!h]
    \centering
    \caption{Number of interventions in runs that simulate different LiDAR frequencies. The tests occurred on the same A and B lanes.}
    \begin{tabular}{ p{1.2cm} p{0.4cm} p{0.4cm} p{0.4cm} p{0.4cm} p{0.4cm} p{0.4cm}}
       \cline{2-7}
        &\multicolumn{6}{ c }{Run (Lane)}\\
        \cline{2-7}
        &\multicolumn{2}{ c }{40~Hz} &\multicolumn{2}{ c }{10~Hz} &\multicolumn{2}{ c }{5~Hz}\\
        \cline{2-7}
        &1(A) &2(B) &1(A) &2(B) &1(A) &2(B)\\
        \hline
        PL+EKF   &1 &2 &1 &2 &1 &2\\
        PL &2 &5 &2 &8 &1 &8\\
        \hline
    \end{tabular}
    \label{tab:variable_lidar}
\end{table}

Twenty eight interventions happened in the sparse rows (C and D in Table \ref{tab:auto_farm_interventions_sparse_cluttered}) with 8 end-of-lane, two operator stops, one bad start and one hardware issue. From the remaining 16, all collisions because of a perception failure, the most significant error sources are heading estimation (8) and converging aspect of the perceived rows (3) because of hanging leaves and fallen stalks occluding the actual rows. Similar to the issue in the gaps (see Fig. \ref{fig:fig_heading_in_gap_ekf_vs_nonekf}), the sparseness in the rows randomly diminished the available readings from the actual row, which raises the possibility of readings from fallen stalks and hanging leaves to be considered as extension of the rows. While PL tests concentrated 14 of these perception-related collisions, PL+EKF had only two (both heading estimation failures).

Finally, the cluttered lanes had twenty six interventions (E and F in Table \ref{tab:auto_farm_interventions_sparse_cluttered}). They had 8 end-of-lane, four bad starts and one hardware issue. The other 13 ended in collision due to a perception error, and again, the most significant source is heading estimation (8), now due to row occlusion because of the presence of high grass and weeds within lane. Similarly, most of the perception-related collisions happened on the PL experiments (10 against 3 in PL+EKF). 

\begin{table}[ht]
  \centering
  \caption{Number of interventions in the sparse with frequent hanging leaves or fallen stalks (C and D) and cluttered with high grass and weeds (E and F) lanes in \autofarm.}
  \begin{tabular}{ p{1.2cm} p{0.4cm} p{0.4cm} p{0.4cm} p{0.4cm}       p{0.4cm} p{0.4cm} p{0.4cm} p{0.4cm}}
     \cline{2-9}
      &\multicolumn{8}{ c }{Run (Lane)}\\
      \cline{2-9}
      &1(C) &2(D) &3(C) &4(D) &5(E) &6(F) &7(E) &8(F)\\
      \hline
      PL+EKF   &2 &1 &1 &2 &1 &3 &1 &2\\
      PL      &6 &9 &4 &3 &4 &5 &4 &6\\
      \hline
  \end{tabular}
  \label{tab:auto_farm_interventions_sparse_cluttered}
\end{table}

%Fifty-four interventions happened in the border rows: 16 end-of-lane, one hardware issue, and 37 collisions. Control, hardware, bad start were responsible for 4 collisions. The other 33 happened due to some error in Perception Subsystem: could not recover from an occlusion (1), wrong distance validation step (6), wrong heading estimation (25), extended occlusion(1), and wrong point picking (2). From the five reasons, the last three either became evident or were intensified because of test characteristics, i.e. significant clutter between rows (external border rows) or scarcity of readings in some regions (internal border rows). 

\subsubsection{\textbf{Controlled Tests} in the production field farm with curve}

\begin{figure}[ht]
  \centering
  \begin{subfigure}{.33\textwidth}
    \includegraphics[height=30mm]{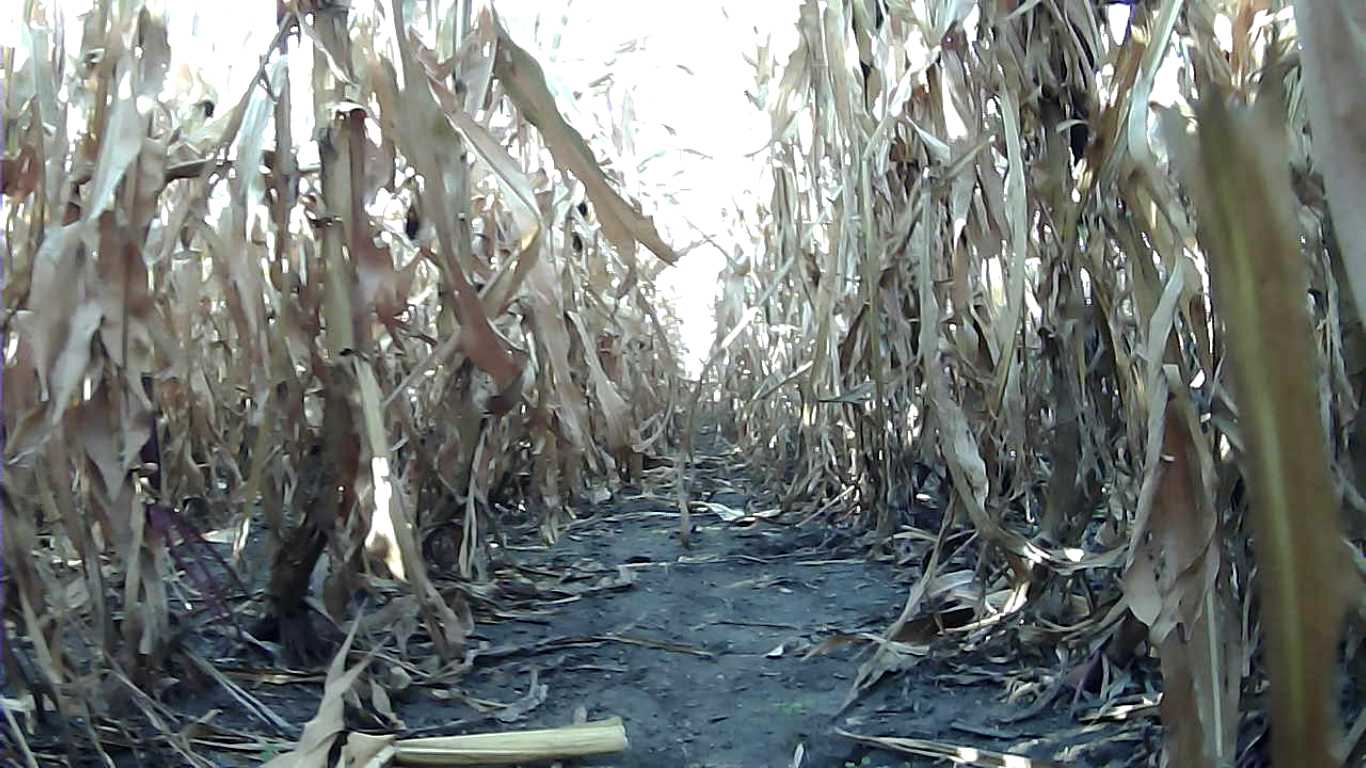}
    \caption{}
    \label{fig:production_field_average_condition}
  \end{subfigure}
  \begin{subfigure}{.32\textwidth}
    \includegraphics[height=30mm]{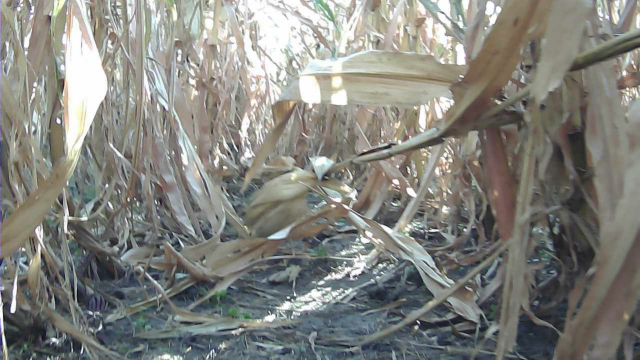}
    \caption{}
    \label{fig:production_field_fallen_stalk}
  \end{subfigure}
  \begin{subfigure}{.33\textwidth}
    \includegraphics[height=40mm]{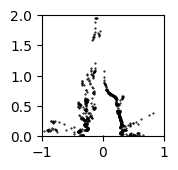} 
    \caption{}
    \label{fig:production_field_fallen_stalk_lidar}
  \end{subfigure}
  \caption{Average condition (a) and an example of fallen stalks and hanging dry leaves (b) and its associated LiDAR scan (c) in the production field corn crop.}
  \label{fig:production_field}
\end{figure}

Additional six runs were performed in a production field corn crop containing a curved section with minimum curvature radius of 8~m (Fig. \ref{fig:production_field_average_condition}). The length of the lanes was around 300~m. Besides the curve, which assesses how well perception performs when rows are not parallel, this field has been left untouched for the season and therefore it possesses all challenges at once: angled stalks (lodging), a high number of dry leaves, fallen plants and ground clutter. The angled stalks increases the possible lane width because the stalks grew outward. The dry leaves may simply block the LiDAR field of view, form an inner "row", form a block in one side that does not allow seeing through or form a converging pattern. The fallen plants may be insurmountable obstacles, cover the row or provide extra clutter. The ground clutter makes the robot unstable, bouncing in all directions. 

 %The angled stalks increases the possible lane width because the stalks grew outward. The dry leaves may simply block the LiDAR, form an inner "row", form a block in one side that does not allow seeing through or form a converging pattern. The fallen plants may be insurmountable obstacles, cover the row or provide extra clutter. The ground clutter makes the robot unstable, bouncing in all directions (curiously, even for PL without EKF, hits in seemingly normal conditions were preceded by some unusual stuff on ground). 
 %Particularly for PL without EKF, occlusions seemed to be highly effective in destabilizing the system. Most PL errors were related to wrongly estimating the heading, a key value to correctly divide readings into right and left sets from where respective lines are derived. This heading error made the opposite set have the top readings of the other side. Therefore, a diagonal line was derived. Another issue was either the validation/line score system that were giving an invalid classification or low score to a seemingly normal line. In summary, production field field tests showed the lack of robustness in the heading estimation step, that rows cannot be treated as parallels (histogram is not a solution for this case) and that the score system is not well defined.  

Seventy two interventions occurred in the six tests, from which 10 will not be considered in this analysis because 2 are hardware issues, 2 are operator manual stop, and 6 are end of lanes. For the remaining 62 interventions, five are bad starts and 46 happened with PL and 11 with PL+EKF. For PL, the most significant error sources were convergence of rows (23), classification of valid side (7) and heading estimation (7). The 11 collisions for PL+EKF occurred because of convergence of rows (4), classification of valid side (3), control not responding to distance error (2), challenging row identification in the LiDAR data (1) and wrong distance estimation (1).

%\miss{There were 72 interventions. They were 27(37.500)\% pl converging rows, 12(16.667)\% pl wrong valid classif, 6(8.333)\% pl heading diverge on occlusion, 6(8.333)\% end of row, 5(6.944)\% bad start, 3(4.167)\% ctrl not responding to offset, 2(2.778)\% pl wrong dist estimate , 2(2.778)\% external interv, 2(2.778)\% locomotion does not follow ctrl, 2(2.778)\% ctrl error due to heading, 2(2.778)\% challenging row identif, 1(1.389)\% pl too close to one side, 1(1.389)\% pl heading error, 1(1.389)\% ctrl priority of heading over cte}

Figures \ref{fig:production_field_fallen_stalk} and \ref{fig:production_field_fallen_stalk_lidar} highlights the issue of having a single scanning plane: there is a horizontal leaf that blocks all readings on right side behind it. Moreover, it could be noted a converging pattern of the rows for $y \in(0, 0.5)$. Because our algorithm assumes that the angle of the best defined line is the heading error, when scan was rotated to make right row parallel to y-axis, left row became highly tilted to right. This prejudiced the point choosing step, which relies on a histogram applied to x-axis to find the 0.05~m bin with most readings. Since right row is already parallel to y-axis, most of its readings were counted in the same bin, but the same does not happen for a diagonal row such as the left whose readings were scattered on several bins. Subsequently, only the readings within the highest count bin and its neighbors are used. Therefore, the derived expression of left row contains only a small section of it.

%\miss{discussion about lane width all together}
%With respect to lane width estimation, the average lane width varied from 0.714 to 0.734 for A with a maximum standard deviation of 0.035~m, and from 0.717 to 0.736 for B with a maximum standard deviation of 0.036~m. At least 98\% fall within 0.1~m error from the average value of respective experiment. The difference from EKF usage becomes evident for the 0.05~m error: while EKF kept at least 94\% for A and 98.5\% for B, the baseline was 83.6\%, 86.8\% for A, and 86.2\% and 86.6\% for B. \miss{Does it make sense to make something similar for all cases?} 

% Figure \ref{fig:comparison_lw_perc_stats} shows that with EKF, lane width dispersion is narrower than PL only as we compare the boxplots for 0.05~m error. But for a 0.1~m error, all percentages are above 95\%.

% \begin{figure}
%     \centering
%     \includegraphics[width=.95\linewidth]{Figures/fig_lw_perc_stats.png}
%     \caption{Two information about estimated lane width are given for each set of experiments: the boxplot shows how the percentage of values below 0.05~m error varies within set and the red cross shows the minimum percentage of estimates below 0.1~m error.}
%     \label{fig:comparison_lw_perc_stats}
% \end{figure}

\subsubsection{Remarks about ground truth}

\begin{figure}[ht]
  \centering
  \includegraphics[width=.7\linewidth]{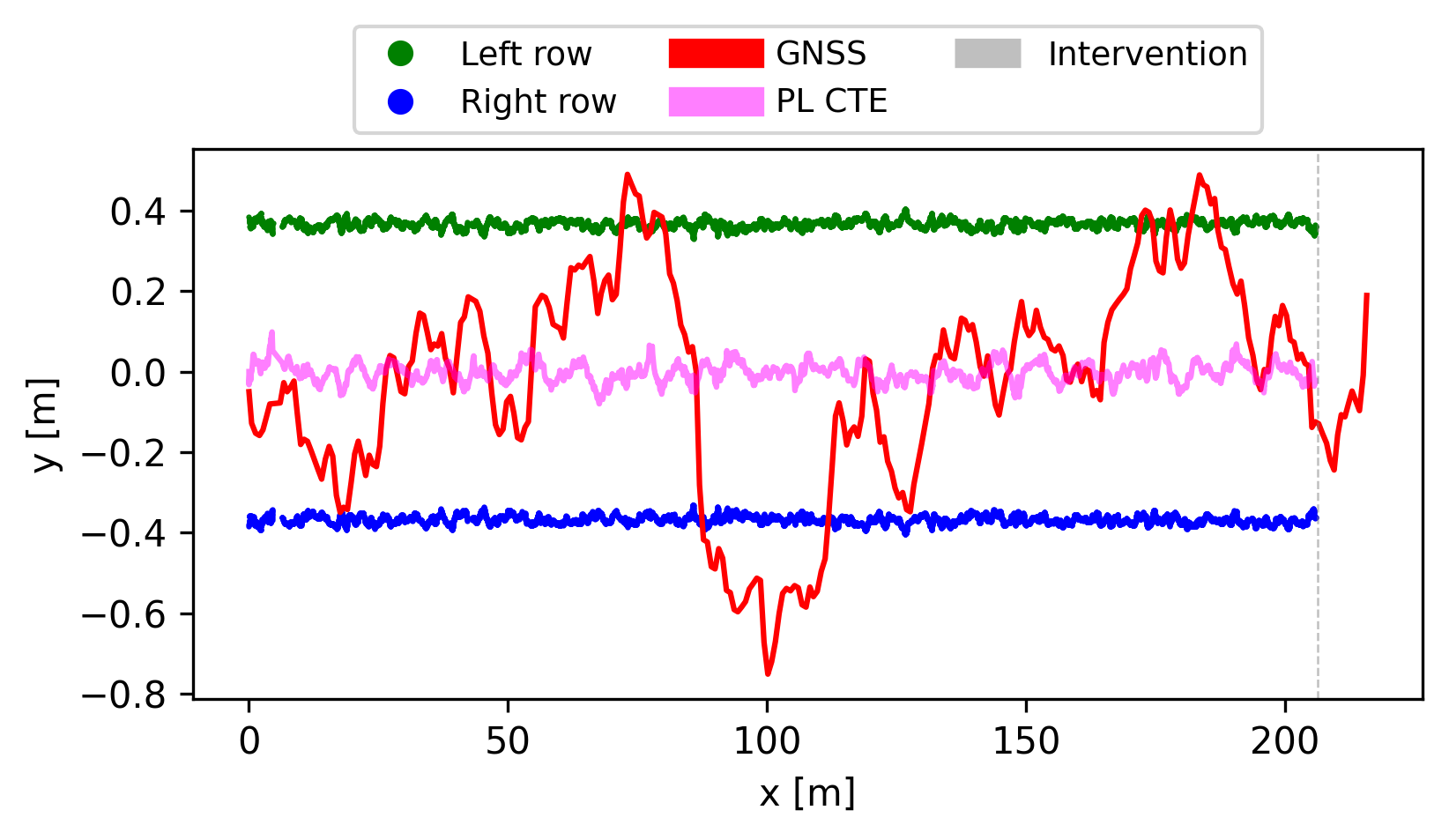}
  \caption{Representation of lane A. The vertical gray line represents the end-of-lane intervention.}
  \label{fig:fig_reconstructed_lane_comparison_straight_collection}
\end{figure}

A representation of \autofarm's lane A is provided on Fig. \ref{fig:fig_reconstructed_lane_comparison_straight_collection}.
%The lighter blue and green lines show the raw estimates, i.e. the values with respect to LiDAR at $(x, 0)$. They do not reflect the rows because they assume that robot only had displacement in x-axis. After accounting for the y-axis displacement (cross track error), the result are the thicker blue and green lines. We may think about these lines as 
The green and blue dots show respectively the left and right lateral rows while magenta line depicts the cross track error. The PL+EKF provides the estimations, which are placed on space by considering the forward speed retrieved by robot's odometry and heading, also estimated by PL+EKF. The GNSS measurements are plotted for comparison (red). Although cross track error and experimental notes indicate that robot stayed within rows and around center of lane, the same cannot be observed from the positioning given by GNSS. 

\begin{figure}[htb]
  \centering
  \includegraphics[width=.7\linewidth]{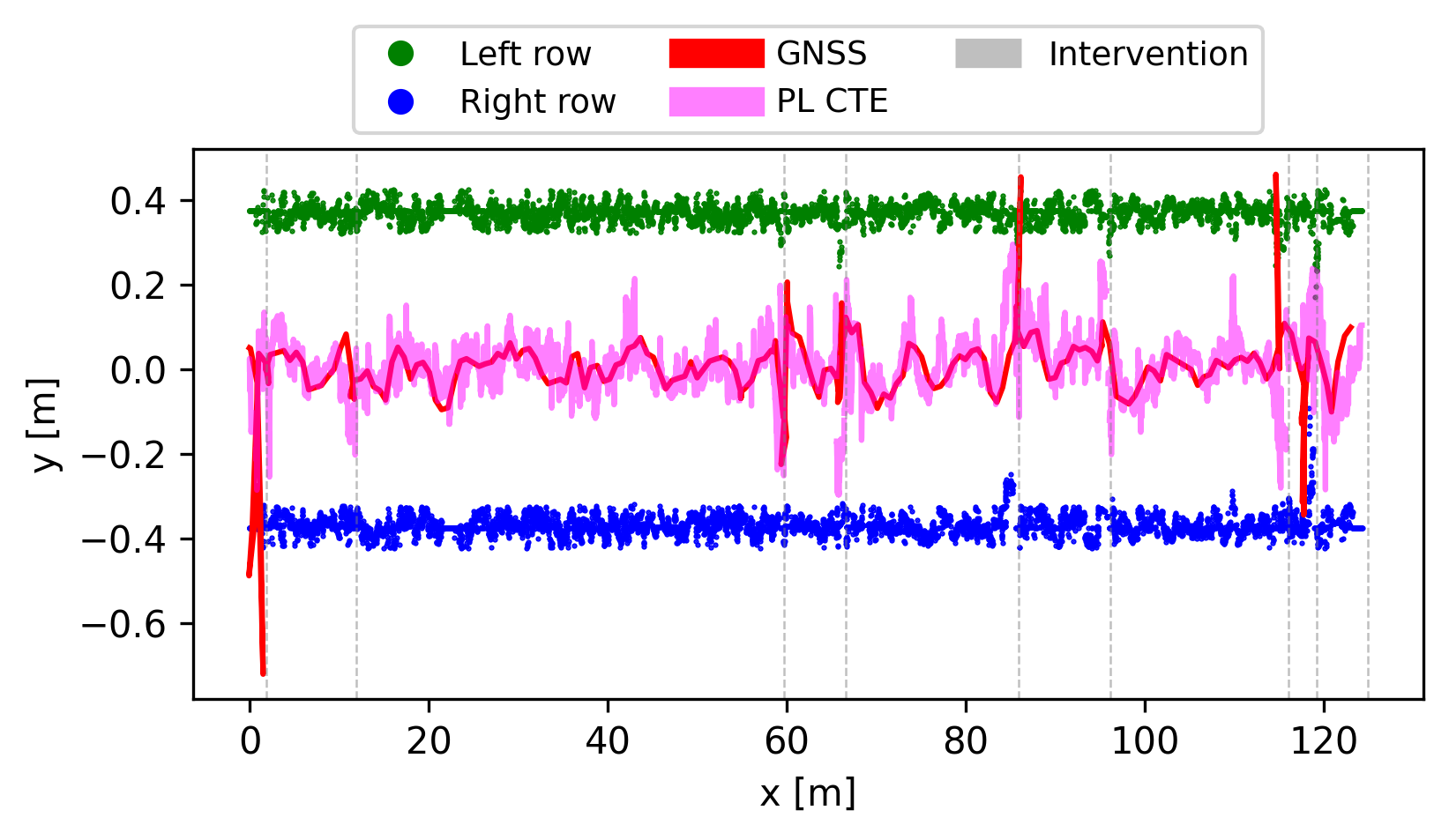}
  \caption{Representation of lane D, an internal border one. Opposed to the experiment in the well behaved lane A in Fig. \ref{fig:fig_reconstructed_lane_comparison_straight_collection}, there were several interventions. }
  \label{fig:fig_reconstructed_lane_comparison_border_rows_internal_nonekf_collection}
\end{figure}

Figure \ref{fig:fig_reconstructed_lane_comparison_border_rows_internal_nonekf_collection} shows a specific occasion on the experiments in the \autofarm\ sparse rows where GNSS achieved the RTK fix quality for most of the time within the lane (D). This was possible because as we can see on Fig. \ref{fig:AutoFarm}, lane A is is the middle of the field while lane D is a border one with some clearance before more crop. Such space may have allowed for the RTK connection a few times. This figure is similar to Fig. \ref{fig:fig_reconstructed_lane_comparison_straight_collection} except that the experiment reports a PL run. The comparison between the green or blue lines show that EKF was able to smooth the estimates and reduce the discontinuities. Although very noise since it is a reconstruction from PL estimates, the cross track error (magenta line) is consistent with the path recorded with GNSS measurements (red line).

\begin{figure}[htb]
  \centering
  \includegraphics[width=0.5\linewidth]{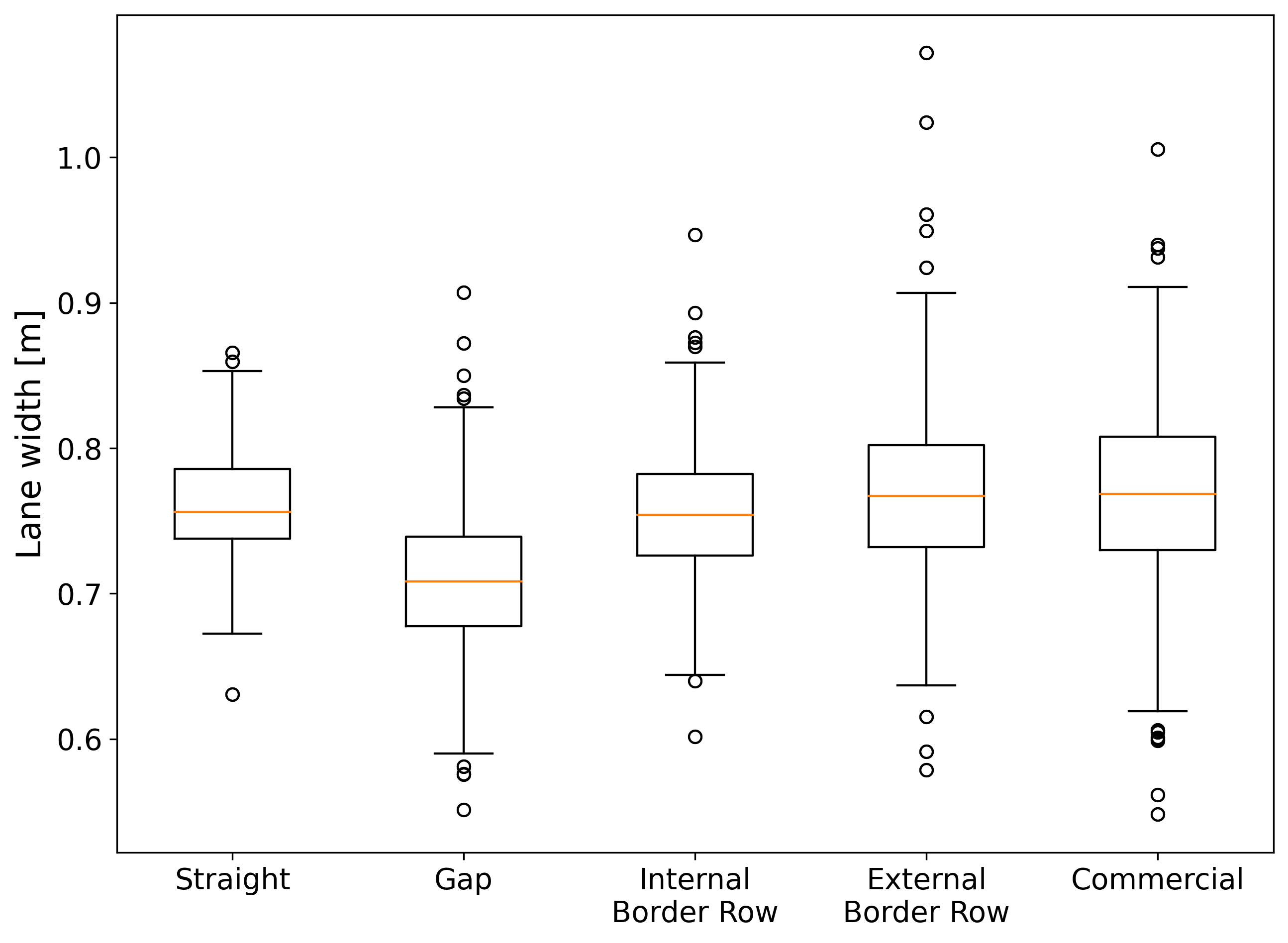}
  \caption{Dispersion of manual ground truth for the controlled scenarios}
  \label{fig:fig_gt_boxplots}
\end{figure}

Manual ground truth was obtained for at least 200 scans for each scenario. Figure \ref{fig:fig_gt_boxplots} shows the dispersion for the manual lane width. Although nominal row spacing is 0.75~m, actual lane width greatly varies and we can expect values from 0.548 to 1.072~m, the most extreme outliers represented by the black circles in the picture.
%although at least 83.884\% should be within 0.1~m difference.

Table \ref{tab:manual_ground_truth} shows more details. Both sides may not be always detectable on a scan (see percentage of valid sides) due to occlusion by leaves, gap or sparseness. With respect to difference between annotated lane width and expected value, the difference is lower in the continuous straight lane as expected (83.246\% within 0.05~m). Such difference is higher on the other scenarios as percentages within 0.05 or 0.1~m get lower. This is caused due to the environment characteristics. Gaps are particularly problematic because they introduce several begins and ends of lane. The begin may be blocked by leaves, weeds, fallen plants and LiDAR may not clearly see ahead. There is a progressive drop in readings as robot approaches the end. These two interferes in the process of finding a line representation for the respective side.  Similarly, sparseness of plants (mostly in internal border rows) make the system more prone to consider eventual cluster of leaves or weeds as part of the row since there are few plants actually forming the row. In the cluttered lanes, there was a great influence of ground unevenness combined with the large amount of clutter along lane. Finally, in the production field crop, besides mentioned issues, lane width consistency is compromised due to corn that grew angled, the curve and higher foliage.

\begin{table}[]
\centering
\caption{Manual ground truth for the controlled scenarios}
\begin{tabular}{cccccc}
\cline{2-6}
& \begin{tabular}[c]{@{}c@{}}Valid \\ left\\ side{[}\%{]}\end{tabular} & \begin{tabular}[c]{@{}c@{}}Valid \\ right\\ side {[}\%{]}\end{tabular} & \begin{tabular}[c]{@{}c@{}}Avg. \\ LW \\ {[}m{]}\end{tabular} & \begin{tabular}[c]{@{}c@{}}LW  \\ 0.05 m \\ {[}\%{]}\end{tabular} & \begin{tabular}[c]{@{}c@{}}LW \\ 0.1 m \\ {[}\%{]}\end{tabular} \\ \hline
Straight & \ 96 & \ 99.500 & \ 0.762 & \ 83.246 & \ 97.906 \\
Gap & \ 96.011 & \ 93.162 & \ 0.708 & \ 53.355 & \ 87.700 \\
\begin{tabular}[c]{@{}c@{}}Internal\\ Border Row\end{tabular} & \ 78.200 & \ 91.600 & \ 0.754 & \ 76.218 & \ 95.129 \\
\begin{tabular}[c]{@{}c@{}}External\\ Border Row\end{tabular} & \ 84.053 & \ 97.010 & \ 0.771 & \ 65.164 & \ 89.344 \\
Production Field & \ 94.340 & \ 81.761 & \ 0.766 & \ 54.959 & \ 83.884 \\ \hline
\end{tabular}
\label{tab:manual_ground_truth}
\end{table}

\subsection{Uncontrolled Tests}

The system has also been tested by third parties, EarthSense Co. partners, in another two locations across US. The datasets have been provided by EarthSense Co. and they will be labelled as ES. The first is a corn research field whose 125~m to 150~m lanes were further divided into 5~m sections with 0.5~m gaps between them. The system was used from August 10th, 2020 to September 21st, 2020. The operator's goal was the collection of visual data, but in this case, for the whole lane. There were 373 trials where 365 lanes were completed with a combination of manual and autonomous mode. The latter was used for 17235.844~m (35\% of total distance). From this set, a subset of 51 experiments (6551~m) was chosen as they met two criteria: 1) Total distance over 100~m; 2) More than 90\% of the traversed distance in autonomous mode
%Such subset contains 138 interventions due to bad start, control, gap or perception issues. %202 interventions (23.821\% of the total 848 for all trials). 
The second is another corn research field whose lanes comprised seventy five 4.8~m sections separated by 0.5~m gaps, which results in 397~m lanes. The row spacing is the standard 0.8~m. The system was used between July 31st, 2020 and August 27th, 2020. Since operator's task was the visual data collection for either the whole lane or specific sections, expected operation distance varied from as low as 4.8~m to 397~m. The 159 runs autonomously covered 28245.670~m.

Because of goals difference for the system's usage, a significant number of the interventions are either end-of-lane (42 and 13, respectively) when it was expected to drive the whole lane or an operator stop (14 and 227, respectively) when a partial data collection was aimed. Table \ref{tab:interv_uncontrolled_tests} summarizes the interventions relevant for this study: Bad start, Control, Gap and Perception.
%where 83 interventions due to bad start (10), control (7), gap (93) or perception issues (28).

\begin{table}[ht]
\centering
\caption{Interventions in \textbf{Uncontrolled Tests}. Percentage with respect to total number of relevant interventions for that location.}
\label{tab:interv_uncontrolled_tests}
\begin{tabular}{cccccc}
\hline
             & Bad start   & Control     & Gap         & Perception  & Total \\ \hline
ES \#1 & 10 (7.2\%)  & 7 (5.1\%)   & 93 (67.4\%) & 28 (20.3\%) & 138   \\ \hline
Es \#2 & 10 (12.1\%) & 13 (15.7\%) & 29 (34.9\%) & 31 (37.3\%) & 83    \\
\end{tabular}
\end{table}

\section{Discussion}

\begin{figure}[ht]
    \centering
    \includegraphics{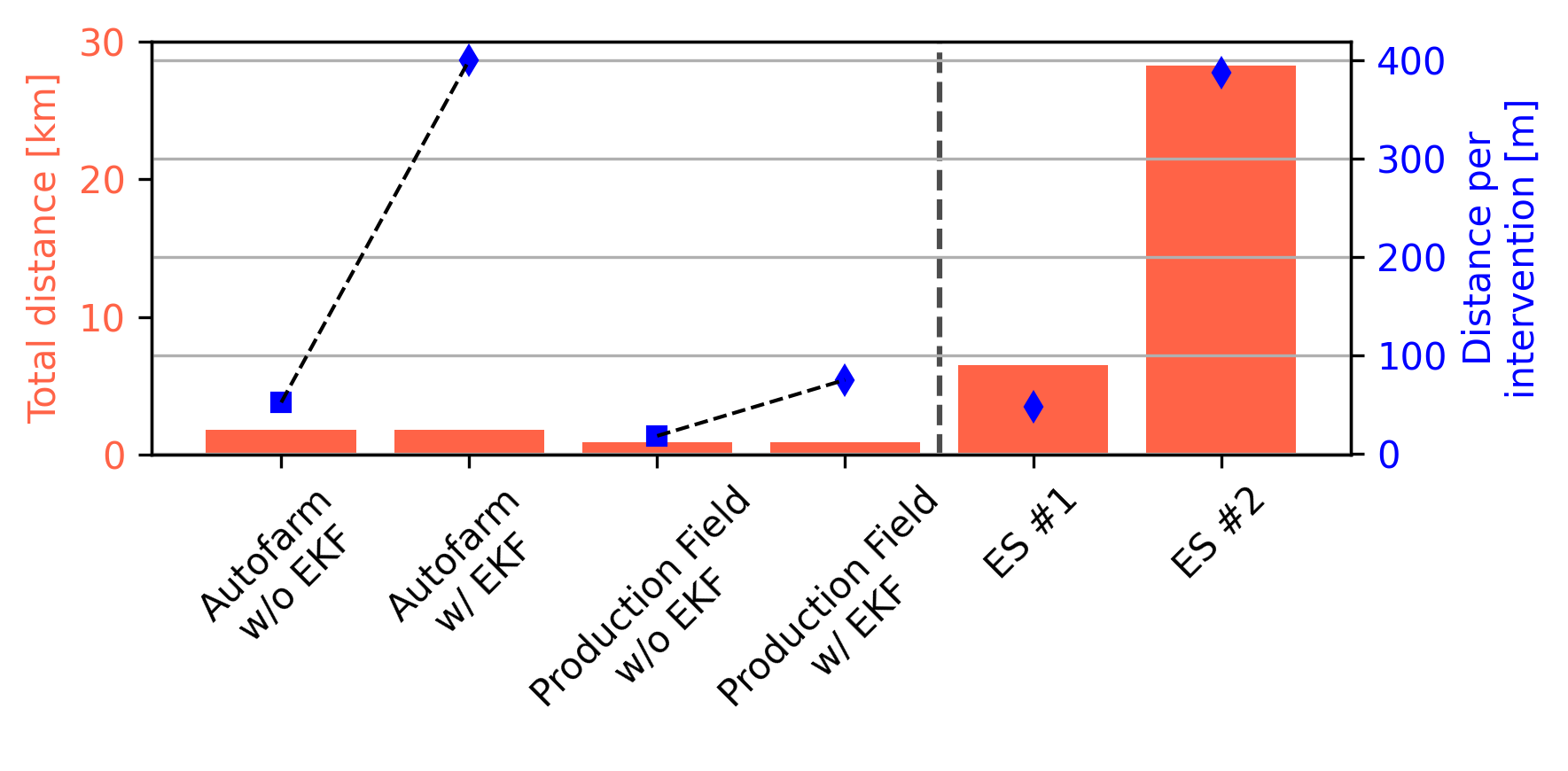}
    \caption{Distance per intervention. Note how \autofarm\ went from 51.6~m/interv without EKF (PL) to 400~m/interv with EKF (PL+EKF). Production field also increased from 16.3~m/interv without EKF to 56.1~m/interv with EKF. Both ES tests were performed with EKF. ES \#1 was heavily affected by gaps and achieved 47.5~m/interv while ES \#2 had similar performance to \autofarm\ with 386.9~m/interv.}
    \label{fig:chart_distance_intervention_paper}
\end{figure}

The \textbf{Controlled Tests} exposed TerraSentia to a variety of scenarios not tested in previous studies. It performed as expected in the well-behaved case with long continuous rows (lane A). The other scenarios ended with a variety of interventions from which we consider relevant the ones caused by a bad start, control, gap or perception error. In this sense, Fig. \ref{fig:chart_distance_intervention_paper} summarizes all experiments with a \textit{distance per intervention} metric, which is defined as the total distance divided by the total number of interventions from the aforementioned four relevant sources. It clearly shows that the EKF use on \autofarm\ and Production Field improved the system's performance. The \textbf{Uncontrolled Tests} used PL+EKF from the start with ES \#2 displaying a similar performance (386.9~m/interv) to \autofarm\ with EKF (400~m/interv) over an extended distance (28~km). Unfortunately, ES \#1 had a major influence of gaps (67.4\% of considered interventions) which greatly reduced the distance per intervention to 47.5~m/interv (similar to Production Field with EKF - 56.1 m/interv). 

In general, we have detected that the perception subsystem error that most leads to collisions is a heading estimation error. To find the lateral rows, Perception Subsystem assumes that $Y_P$ axis (see Fig. \ref{fig:frames}) divides readings into two regions: one to the left and another to the right. Subsequently, the lateral distance estimate is computed from the respective side set. When heading is wrong, e.g. a $\phi = 0$ in Fig. \ref{fig:frames} and $Y_P$ collapses to $Y_R$, we can expect that top right readings become part of the left set. With this, the left side estimate will point towards right and robot will drive away from left side instead of going towards it to return the robot to the lane center. It should be noted that the scan in Fig. \ref{fig:frames} is ideal while field conditions that lead to this error include partial/complete occlusion, leaves that completely block the stem detection, local convergence of rows, neighbor rows and weeds.

The production field experiments also showed that a convergence of rows affects the lateral distance estimation. Such convergence, either originated from the curve or perceived due to leaves, fallen stalks or weeds occluding the actual rows, breaks the PL assumption that rows are mostly parallel for the next meters \cite{Higuti2018UnderNavigation}. Since PL relies on both pre-rotation of LiDAR data to attempt having at least one of the rows parallel to robot's longitudinal axis and subsequent application of an histogram to detect the rows, the lack of parallelism greatly induces errors in estimation. This may be addressed by adding a classification step to remove unwanted readings from objects other than the stalks and also reformulating PL to consider curved/converging rows.

From Table \ref{tab:interv_uncontrolled_tests}, we can also see how gaps have been highly influential on preventing robot to finish its task to completely traverse lanes. Indeed, a large portion (67.4\%) of the relevant interventions for Location \#1 is gap. Since the core Perception Subsystem is an estimator of lateral distances while within rows, gaps are not addressed and these cases and even break another PL's assumption that robot is dealing with a long and continuous corridor. Therefore, unlike before that changes were hardly expected throughout the lane, a local map becomes attractive to deal with the question of traversing gaps. 

\section{Conclusions}

This work reports 50.88~km field testing of an autonomous solution for under canopy navigation in corn crops. On its \textbf{Controlled Tests}, it showed that distance per intervention is significantly improved when core perception subsystem (PL) is enhanced by the proposed EKF. Such performance is corroborated by the field results from \textbf{Uncontrolled Tests}. The intervention analysis has shown that three major problems remain. Heading estimation and convergence of rows are present for lanes with frequent objects such as hanging leaves, fallen stalks, high grass and weeds, all capable of occluding the actual rows to be followed. The convergence problem also indicates that curves requires further attention. Finally, the most influential source of intervention was the presence of gaps, which greatly affected ES \#1 performance. All of them break the assumptions once devised to constrain the challenge of navigating in corn crop using a LiDAR sensor to make it feasible in real world. While such questions remain open for a robust autonomous navigation, the presented PL+EKF already relieves the robot operator of the tedious task of manually driving the robot to collect field data.   

%\begin{figure}
%\begin{tabular}{cc}
%  \includegraphics[width=40mm]{Figures/CADterra2020.JPG} &  % \includegraphics[width=40mm]{Figures/terra2020.jpg} \\
%(a) & (b) \\[6pt]
%\end{tabular}
%\caption{(a) }
%\label{fig:scenarios}
%\end{figure}

\subsubsection*{Acknowledgments}

The work was partially supported by Sao Paulo Research Foundation (FAPESP) grant number 2018/10894-2. The authors thank EarthSense support with TerraSentia robots and field data. The authors also thank Jacob Washburn, Plant Genetics Researcher at University of Missouri, for performing part of the \textbf{Uncontrolled Tests}.  We also would like to thank DASLAB members (Sri Vuppala Theja, Nolan Replogle, Billy Doherty and Rajan Aggarwal) for the technical support in the deployment of the proposed system.

\bibliographystyle{apalike}
%\bibliography{frExampleRefs, all_refs, references}
\bibliography{frExample}

\end{document}